\documentclass[letterpaper, 10 pt, conference]{ieeeconf}  

\IEEEoverridecommandlockouts                              

\usepackage{cite}  
\usepackage{graphicx} 
\usepackage{amssymb}
\usepackage{amsmath}
\usepackage{bm}
\usepackage{xcolor}
\usepackage{siunitx}
\usepackage{subcaption}
\usepackage{float}
\usepackage{placeins}
\usepackage{url}
\usepackage{xcolor}
\usepackage{soul}
\definecolor{authorHL}{RGB}{135, 206, 235}
\definecolor{rev10HL}{RGB}{180, 255, 180}
\definecolor{rev2HL}{RGB}{255, 255, 50}
\definecolor{fer}{RGB}{233, 50, 50}
\renewcommand{\hl}[1]{#1} 
\title{\LARGE \bf
VBT-MPC: Vision-Based Tactile MPC for Contour Following
}
\newcommand\scalemath[2]{\scalebox{#1}{\mbox{\ensuremath{\displaystyle #2}}}}

\newcommand{\world}{\mathcal{B}}
\newcommand{\camera}{\mathcal{C}}
\newcommand{\image}{\mathcal{I}}

\newcommand{\camerasmall}{\scalemath{0.6}{\mathcal{C}}}
\newcommand{\imagesmall}{\scalemath{0.6}{\mathcal{I}}}

\newcommand{\point}{\scalemath{0.5}{P}}
\newcommand{\edge}{\scalemath{0.5}{E}}
\newcommand{\pointone}{\scalemath{0.5}{P_1}}
\newcommand{\pointtwo}{\scalemath{0.5}{P_2}}

\author{Edison Velasco-Sanchez$^{1\dagger}$, Luis F. Recalde$^{2\dagger}$, Guanrui Li$^{2}$, and Pablo Gil$^{1}$
\thanks{*This work was supported by the Interreg-VI Sudoe and European Regional Development Funds through the REMAIN Project under Grant S1/1.1/E0111.}
\thanks{$^\dagger$These authors contributed equally.}
\thanks{$^{1}$ The authors are with AUROVA Lab, Computer Science Research
Institute, University of Alicante, 03690 Alicante, Spain. {\tt\footnotesize email: edison.velasco@ua.es, pablo.gil@ua.es}}%
\thanks{$^{2}$
The authors are with the Worcester Polytechnic Institute, Robotics Engineering, Worcester, MA 01609, USA. {\tt\footnotesize email: lfrecalde@wpi.edu, gli7@wpi.edu}.}
\thanks{This article has been accepted for publication in IEEE Robotics and Automation Letters. 
This is a preprint version. 
\copyright~2026 IEEE. Personal use of this material is permitted. 
Permission from IEEE must be obtained for all other uses.}
\thanks{Additional materials are available at:}
\thanks{\protect\url{https://aurova-projects.github.io/VBT-MPC/}}
}

\begin{document}

\maketitle

\begin{abstract}

Tactile sensing\sethlcolor{rev10HL}\hl{ plays a key role in robotic manipulation, particularly in tasks like surface inspection. Successful execution requires maintaining contact while accurately tracking object contours. In this work, we propose a Vision-Based Tactile Model Predictive Control (VBT-MPC) framework for robotic contour following using a Vision-Based Tactile Sensor (VBTS) mounted in an eye-in-hand configuration. The proposed controller operates directly in contour features space, thereby avoiding the need for separate pose-estimation modules or complex force-control architectures. We further compare our VBT-MPC with visual-servoing strategies adapted to tactile features, and evaluate contour tracking on objects with diverse geometries and materials in both simulation and real-world experiments.}

\end{abstract}

\section{INTRODUCTION}

Touch sensing provides significant advantages for robotic manipulation tasks that cannot be obtained from vision alone. Tactile sensors enable robots to detect contact forces, pressure, and texture; providing a more refined description of the physical interaction with the environment \cite{2015tactile,lepora2026tactile}. \sethlcolor{rev10HL}\hl{This capability is especially important in manipulation} tasks that require controlled exploration of the surfaces of objects. 



Vision-Based Tactile Sensors (VBTS) have gained attention due to their high spatial resolution and relatively low cost compared to resistive, capacitive, magnetic, or piezoelectric tactile arrays \cite{Zhang2022,He2025}. \sethlcolor{rev10HL}\hl{VBTS captures pixel-level contact information, 
by imaging the deformation of a compliant elastomer with an internal camera, and can be broadly categorized into two approaches: marker-based and markerless}\cite{Li2025}. Marker-based sensors track the motion of internal markers embedded in the compliant medium, whereas marker-less designs infer contact information by analyzing changes in the surface appearance caused by the deformation. The resulting tactile images contain rich information from which higher-level features, such as contact position, local curvature, or contours, can be extracted and used to control the robot in active perception tasks \cite{Luo2025review}.



Tactile servoing methods are commonly categorized into \sethlcolor{rev10HL}\hl{Image-Based Tactile Servoing (IBTS) and Pose-Based Tactile Servoing (PBTS)}~\cite{Lepora2021}. \hl{PBTS regulates the sensor pose using learned models that map tactile images to local poses.} However, this approach requires supervised training with ground truth poses and estimation models~\cite{Lepora2021, LLoyd2024}. \hl{In contrast, IBTS defines the control objective directly in the feature space of tactile images}~\cite{sikka2005tactile, KAPPASSOV2020}, making it conceptually simpler and independent of pose-estimation models. \hl{However, existing IBTS approaches for contour-following rely on additional force-regulation controllers to maintain succesful interaction, which increases the overall control complexity.}

 \hl{In this work, we propose a Vision-Based Tactile Model Predictive Control (VBT-MPC) framework shown in Fig.}~\ref{fig:mpc_pipeline}, \hl{which regulates contour features directly from marker-less tactile images without supervised pose estimation or cascaded force controllers, while enforcing contact, tactile features within the field of view (FoV), and input constraints.} In summary, the main contributions of this paper are as follows:

\begin{figure}[t]
    \centering
     \includegraphics[width=\columnwidth]{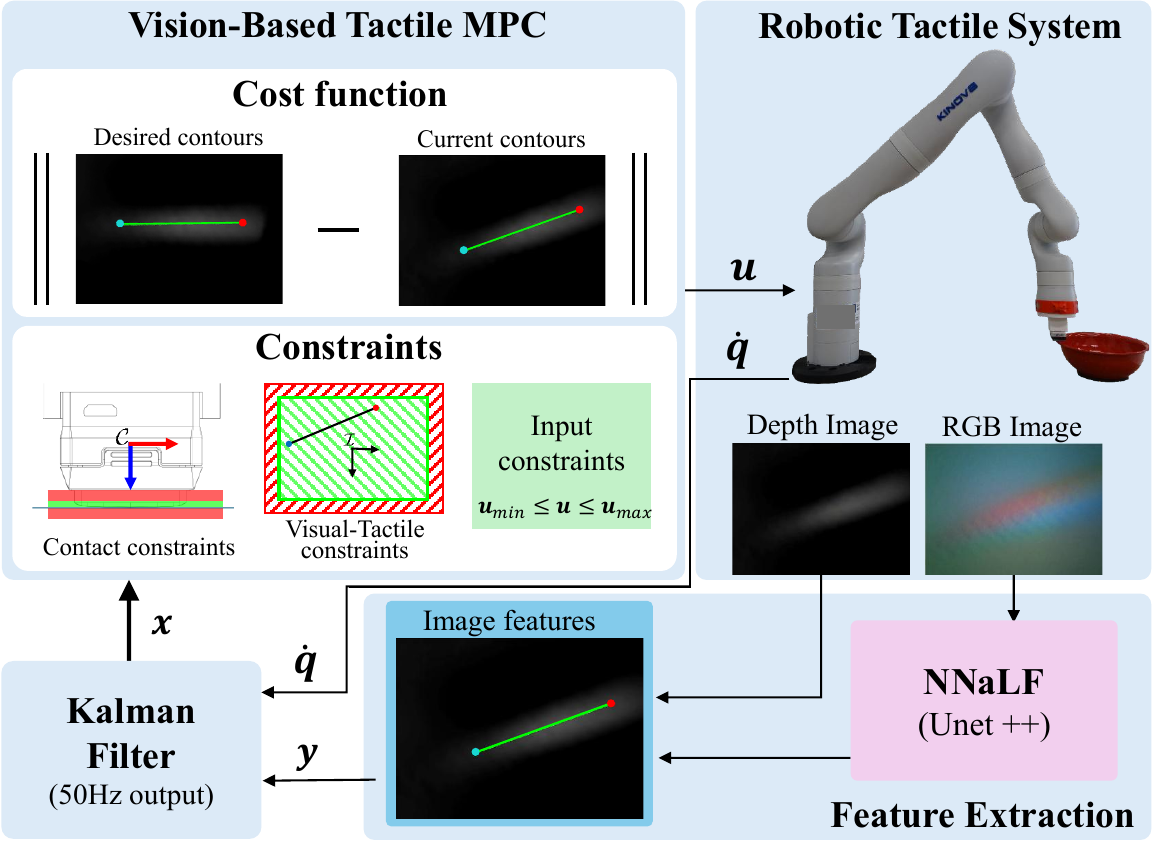}
    \caption{Proposed VBT-MPC for contour following by touch.}
    \label{fig:mpc_pipeline}
\end{figure}

\begin{itemize}
    \item A 
    VBT-MPC framework for contour following that maximizes forward velocity while enforcing touch-maintenance and perception constraints arising from tactile feature extraction. Our framework  operates directly on the contour feature space obtained from marker-less VBTS measurements.
    
    \item A contour feature extraction pipeline that integrates a segmentation network, line fitting, and an Extended Kalman Filter (EKF) to estimate and  over-sample contour features. We quantitatively evaluated this pipeline against three contour-extraction baselines on tactile images acquired from VBTS.

    \item An experimental validation of the proposed VBT-MPC framework both in simulation and in the real-world. The proposed controller is compared with couple and decoupled visual servoing strategies adapted to tactile features for contour-following tasks. The experiments involve 3D-printed and real-world objects, revealing improved tracking performance of our method.
\end{itemize}

The remainder of the paper is structured as follows. Section~\ref{sec:related_work} introduces VBTS for perception and motion control. Section~\ref{sec:method} details the proposed VBT-MPC controller and the contour feature extraction pipeline. Section~\ref{sec:base_method} presents the baseline methods used to compare our approach. Section~\ref{sec:experimentation} reports the experimental results. Finally, Section~\ref{sec:conclusions} summarizes the findings and future research directions.
\section{RELATED WORKS}
\label{sec:related_work}


Early image-based tactile servo control approaches treated tactile arrays as pressure images. For instance,~\cite{Qiang2013} proposed a control scheme based on features extracted directly from the pressure distribution. This idea was later extended by \cite{Li2015}, which introduced an IBTS framework that used similar pressure-image features to perform force control, rolling, and sliding. More recently,~\cite{Kappassov2022} augmented this representation to include information from the center of the pressure image, to address more complex interaction tasks. Despite these advances, such methods rely on pressure-derived descriptors that are subsequently projected into robot space, rather than exploiting contour-based features directly within the tactile image domain, as proposed in this work.


With the advent of VBTS, several works have explored contour perception and pose estimation from tactile images. \cite{Lepora2019} proposed a deep convolutional neural network capable of detecting contours in tactile images acquired with TacTip, a VBTS with markers. This capability allows a robot to plan contact points along the contours of the object and perform a PBTS methodology \cite{Lepora2021}. More recently, in \cite{LLoyd2024}, the authors introduced a contact pose estimator that combined a convolutional neural network with a Bayesian filtering framework to infer the pose from VBTS measurements. The resulting contact pose was then used in a PBTS control scheme. Finally, \cite{Aquilina2024} extended this line of work to 2D contour following in dynamic environments by proposing a shear controller that minimized tactile shear deformation in a marker-based VBTS. 

In summary, tactile servoing approaches based on tactile arrays typically encode feedback as force features that are projected into the robot space, whereas VBTS-based methods predominantly follow PBTS formulations in which deep neural networks estimate the contact pose from marker-based tactile images before control is applied. In both cases, the control objective is expressed in terms of end-effector pose or contact forces, rather than directly in the tactile image domain.
In contrast, our work performs contour tracking directly in the tactile image space using contour features extracted from a marker-less VBTS within an IBTS framework. We introduce a MPC formulation for contour following that incorporates touch perception, and input constraints through geometric contour parameters derived from segmented tactile images, eliminating the need for pose estimation models and cascaded force controllers.

\section{METHOD FOR CONTOUR-FOLLOWING USING TACTILE SENSING}
\label{sec:method}
In this section, we introduce the mathematical representation of contour features extracted from marker-less VBTS images. These features form the basis of our Vision-Based Tactile MPC framework and contour–feature extraction pipeline. 

\subsection{Tactile Features Model}
\label{sec:tactile_features}

\begin{figure}[t]
    \centering
     \includegraphics[width=1\columnwidth]{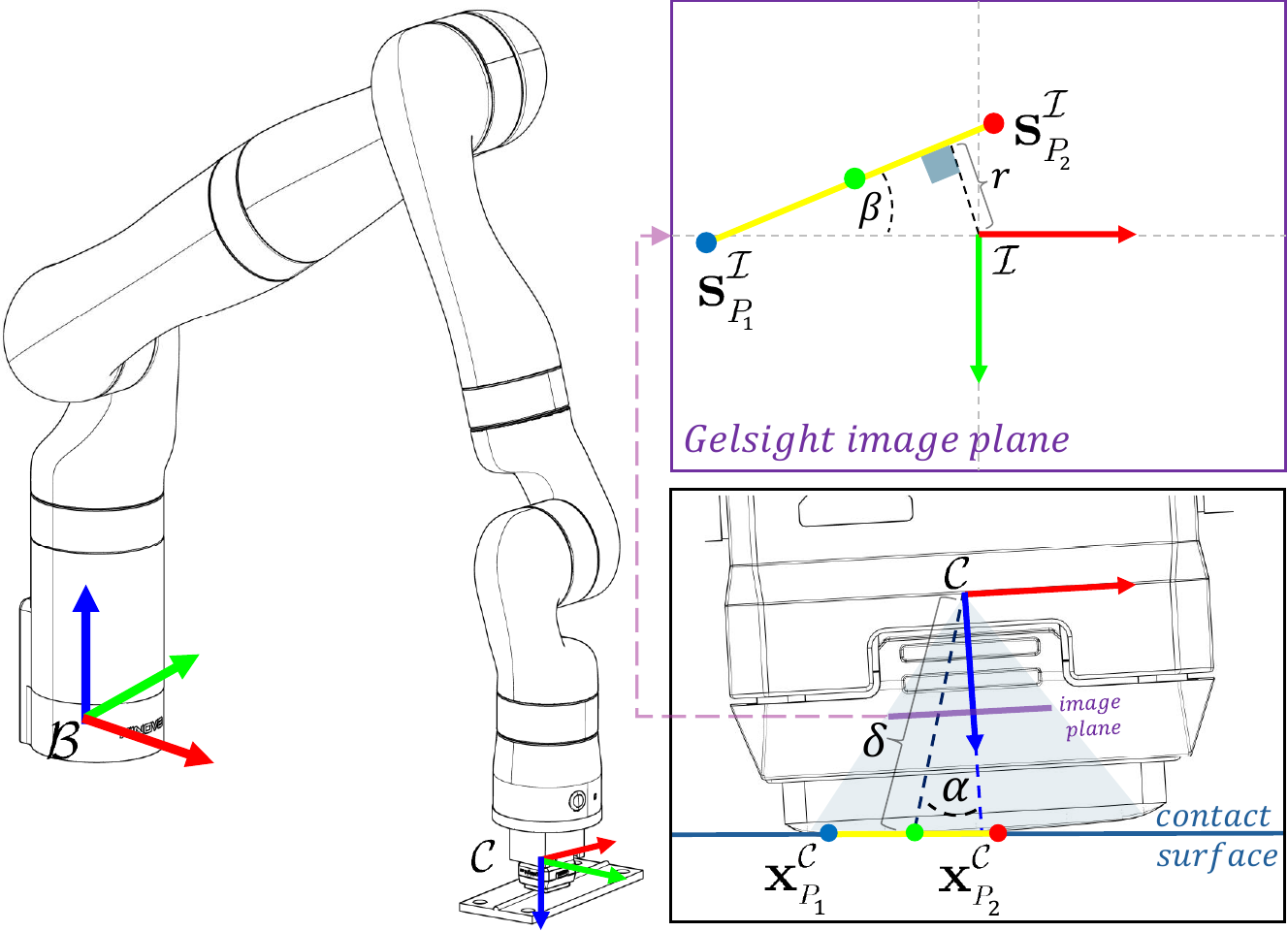}
    \caption{Coordinate frames and geometric features used in our VBT-MPC framework.}
    \label{fig:features_parametrization}
\end{figure}

A tactile robotic system enables the guidance of a manipulator using tactile perception. In our setup, we adopt an eye-in-hand configuration in which the robot end-effector is equipped with a VBTS. \sethlcolor{rev10HL}\hl{We consider three coordinate frames: the robot base frame $\world$, the camera frame $\camera$, and the image frame $\image$, as illustrated in }Fig.~\ref{fig:features_parametrization}.

As the end-effector interacts with the contact surface, the VBTS provides a set of tactile features detected directly in the camera frame. Let $\mathbf{x}_{\point}^{\camerasmall} \in \mathbb{R}^{3}$ denotes the coordinates of a raw tactile feature, 
its projection onto the image frame $\image$ can be written as $\mathbf{s}_{\point}^{\imagesmall} \in \mathbb{R}^{3}$, and it is defined as follows:
\begin{align}
\mathbf{s}_{\point}^{\imagesmall} &= \mathcal{P}\!\left(\mathbf{x}_{\point}^{\camerasmall}, \mathbf{K}\right), \\
\dot{\mathbf{s}}_{\point}^{\imagesmall} &= 
\mathbf{J}_{\point}\!\left(\mathbf{s}_{\point}^{\imagesmall}\right)\,\boldsymbol{\upsilon}^{\camerasmall},
\label{eq:image_projection}
\end{align}
where $\mathcal{P}(\cdot):\mathbb{R}^{3} \rightarrow \mathbb{R}^{3}$ denotes the projection map, $\mathbf{K}$ is the intrinsic calibration matrix of the VBTS, and $\mathbf{J}_{\point} \in \mathbb{R}^{3 \times 6}$ is the corresponding interaction matrix. Being 
$\boldsymbol{\upsilon}^{\camerasmall} = \begin{bmatrix} \mathbf{v} & \boldsymbol{\omega} \end{bmatrix}^{\top} \in \mathbb{R}^{6}$ a representation of the end-effector twist, with $\mathbf{v} 
$ and $\boldsymbol{\omega} 
$ denoting the linear and angular velocity components, respectively. 

To formulate our contour-following controller, we introduce an additional set of features denoted by $\boldsymbol{\xi}^{\imagesmall} \in \mathbb{R}^{4}$, which represent a contour feature vector.
This vector is defined from two tactile features $\mathbf{s}_{\pointone}^{\imagesmall}$ and $\mathbf{s}_{\pointtwo}^{\imagesmall}$ as follows:
\begin{align}
\boldsymbol{\xi}^{\imagesmall} &= \mathcal{E}\!\left(\mathbf{s}_{\pointone}^{\imagesmall}, \mathbf{s}_{\pointtwo}^{\imagesmall}, \mathbf{K}\right), \\
\dot{\boldsymbol{\xi}}^{\imagesmall} &= 
\mathbf{J}_{\edge}\!\left(\mathbf{s}_{\pointone}^{\imagesmall}, \mathbf{s}_{\pointtwo}^{\imagesmall}\right)\,\boldsymbol{\upsilon}^{\camerasmall},
\label{eq:edge_features}
\end{align}
 where $\mathcal{E}(\cdot): \mathbb{R}^{6} \rightarrow \mathbb{R}^{4}$ denotes the nonlinear operator that maps tactile features to contour features, 
and $\mathbf{J}_{\edge} \in \mathbb{R}^{4 \times 6}$ is the corresponding Jacobian matrix. Being $ \boldsymbol{\xi}^{\imagesmall} = 
\begin{bmatrix}
     r & \beta  & \alpha & \delta
\end{bmatrix}^{\!\top}$ \sethlcolor{rev10HL}\hl{where $(r,\beta)$ encode the orthogonal distance $~\mathrm{[mm]}$ and orientation $~\mathrm{[rad]}$ of the detected contour in the tactile image plane, $\alpha$ denotes the tangential orientation $~\mathrm{[rad]}$ of the sensor with respect to the local contact surface, and $\delta$ represents the local surface deformation $~\mathrm{[mm]}$ estimated from the reconstructed tactile depth image} \cite{Yuan2017,wang2021gelsight}. \hl{Since the tactile RGB image and the reconstructed depth image are spatially aligned, the value of $\delta$ is obtained through pixel-wise correspondence between the detected contour location in the image plane and its associated depth value. The geometric interpretation of these features and the involved coordinate frames are illustrated in Fig}.~\ref{fig:features_parametrization}.


With a slight abuse of notation, the tactile robotic system considering the visual tactile features 
presented in \eqref{eq:image_projection} and \eqref{eq:edge_features}, can be compactly written as
\begin{equation}
\begin{aligned}
\dot{\mathbf{x}} = \mathbf{f}(\mathbf{x}, \mathbf{u}) 
&\triangleq 
\begin{bmatrix}
\mathbf{J}_{\pointone}\!\left(\mathbf{s}_{\pointone}^{\imagesmall}\right)\, \boldsymbol{\upsilon}^{\camerasmall} \\[1mm]
\mathbf{J}_{\pointtwo}\!\left(\mathbf{s}_{\pointtwo}^{\imagesmall}\right)\, \boldsymbol{\upsilon}^{\camerasmall} \\[1mm]
\mathbf{J}_{\edge}\!\left(\mathbf{s}_{\pointone}^{\imagesmall}, \mathbf{s}_{\pointtwo}^{\imagesmall}\right)\, \boldsymbol{\upsilon}^{\camerasmall}
\end{bmatrix},
\end{aligned}
\label{eq:compact_system}
\end{equation}
where $\mathbf{x} = \begin{bmatrix}
    \mathbf{s}_{\pointone}^{\imagesmall} &\mathbf{s}_{\pointtwo}^{\imagesmall}&\boldsymbol{\xi}^{\imagesmall}
\end{bmatrix}^{\top}\in \mathbb{R}^{10}$ collects the pixel coordinates of the two raw tactile features and the additional tactile features computed as contour's edge, 
where \mbox{$\mathbf{u} = \boldsymbol{\upsilon}^{\camerasmall}\in \mathbb{R}^{6}$} denotes the twist of the end-effector.

\subsection{Vision-Based Tactile MPC}
In this section, we introduce our VBT-MPC framework for contour following with robotic manipulators illustrated in 
Fig.~\ref{fig:mpc_pipeline}. The controller is designed to achieve contour tracking while maximizing forward velocity and simultaneously enforcing contact constraints, perception constraints arising from tactile feature extraction, and input constraints associated with the end-effector velocity. 

MPC strategies compute a sequence of future 
states and control inputs that minimize a cost function 
over a finite prediction horizon, while explicitly incorporating the 
system model and constraints. 
A general MPC formulation can be expressed as follows:
\begin{equation}\label{eq:nmpc}
\begin{aligned}
\underset{
    \substack{\mathbf{x}_1,\ldots,\mathbf{x}_N,\\
              \mathbf{u}_1,\ldots,\mathbf{u}_{N-1}}
}{\arg\min}\quad 
& \; \ell_N(\mathbf{x}_N) 
  + \sum_{k=1}^{N-1} \ell(\mathbf{x}_k, \mathbf{u}_k) \\[1mm]
\text{subject to}\quad
& \mathbf{x}_{k+1} = \mathbf{f}(\mathbf{x}_k,\mathbf{u}_k), 
    \quad k = 1,\ldots, N-1, \\
& \mathbf{g}(\mathbf{x}_k,\mathbf{u}_k) \le 0, 
    \quad k = 1,\ldots, N-1,
\end{aligned}
\end{equation}
where $\ell_N(\mathbf{x}_N)$ and $\ell(\mathbf{x}_k, \mathbf{u}_k)$ denote 
terminal and running cost functions, respectively. 
The constraint $\mathbf{x}_{k+1} = \mathbf{f}(\mathbf{x}_k,\mathbf{u}_k)$ 
represents the dynamics of the discrete-time nonlinear system, while 
$\mathbf{g}(\mathbf{x}_k,\mathbf{u}_k)$ encodes system constraints.

To allow the tactile robotic system to follow the contour, we define an error between the desired contour features $\boldsymbol{\xi}^{\imagesmall}_{d}  =\begin{bmatrix}
r_d & \beta_d & \alpha_d & \delta_d
\end{bmatrix}^{\top}$ and the current features extracted from the sensor $\boldsymbol{\xi}^{\imagesmall}$. The contour error is given by
\begin{equation}
\boldsymbol{\xi}^{\imagesmall}_{e}
=\boldsymbol{\xi}^{\imagesmall}_{d} - \boldsymbol{\xi}^{\imagesmall},
\label{eq:edge_error}
\end{equation}
where $\boldsymbol{\xi}^{\imagesmall}_{e} = \begin{bmatrix}
r_e & \beta_e & \alpha_e & \delta_e
\end{bmatrix}^{\top}$ denotes the deviation of the contour’s centerline from its desired values.

Additionally, to enable the system to advance along the contour at the maximum feasible velocity in regions of low curvature, we define a twist error between the desired twist 
$
\mathbf{u}_d =
\begin{bmatrix}
\mathbf{v}_d & \boldsymbol{\omega}_d
\end{bmatrix}^{\top}
$ and the current twist:
\begin{equation}
\mathbf{u}_e = \mathbf{u}_d - \mathbf{u}.
\label{eq:control_error}
\end{equation}
The desired linear velocity $\mathbf{v}_d = \begin{bmatrix} v_{xd} & v_{yd} & v_{zd} \end{bmatrix}^{\top}$ in $\mathbf{u}_d= \begin{bmatrix} \mathbf{v}_d & \boldsymbol{\omega}_d \end{bmatrix}^{\top}$ is dynamically updated according to the contour error $\boldsymbol{\xi}^{\imagesmall}_{e}$. The forward velocity component is defined as
\begin{equation}
v_{xd} = \frac{v_{x,{\max}}}{1 + \| \boldsymbol{\xi}^{\imagesmall}_{e} \|_{\mathbf{W}}^{2}},
\label{eq:max_vx}
\end{equation}
where $v_{x,{\max}}$ is the maximum velocity along the contour and 
$\mathbf{W} \succ 0$ is a weighting matrix that normalizes the contribution of the contour error, and the operator
$\| \cdot \|_{\mathbf{W}}^{2}$ denotes the squared Euclidean norm.

This dynamic formulation allows the end-effector to move rapidly along low-curvature sections while decreasing the velocity in regions with larger deviations.
Using the error terms defined above, we now construct the running and terminal cost functions for our VBT-MPC framework over a finite horizon $N$
 \begin{equation}\label{eq:cost_function}
\begin{split}
\ell(\boldsymbol{\xi}^{\imagesmall}_{eN}) + \sum_{k=1}^{N-1}\ell(\boldsymbol{\xi}^{\imagesmall}_{ek}) + \ell_{u}(\mathbf{u}_{ek})\,,
\end{split}
\end{equation}
with
\begin{equation}
  \ell(\boldsymbol{\xi}^{\imagesmall}_{e}) = \Vert 
\boldsymbol{\xi}^{\imagesmall}_{e}
  \Vert^{2}_{\mathbf{Q}},\,\,\, \ell_{u}({\mathbf{u}}_e) = \Vert  {\mathbf{u}}_e\Vert^{2}_{\mathbf{R}}.
\end{equation}
The matrices $\mathbf{Q}\succ 0$ and $\mathbf{R}\succ 0$ weight the contour feature and twist vector errors.


We then approximate the tactile robotic system using a discrete form of 
\eqref{eq:compact_system}, constructed with the fourth-order Runge-Kutta method. This can be expressed as
\begin{equation}\label{eq:campact_discrete}
   {{\mathbf{x}}}_{k+1} = \mathbf{f}_{RK4}({\mathbf{f}}({\mathbf{x}}_k,{\mathbf{u}}_k), dt)\,,
\end{equation} 
where $dt$ denotes the sampling interval.

Due to the nature of the VBTS, tactile features are available only when the sensor is in contact with a surface. To avoid the loss of tactile features, we introduce constraints that enforce contact and ensure that the extracted features remain within the FoV of the sensors. In addition, smooth control input constraints are imposed. These conditions can be compactly expressed as
\begin{equation}\label{eq:conditions}
  \mathbf{g}({\mathbf{x}}, {\mathbf{u}}) \leq 0,\,\,\,  
\mathbf{g}({\mathbf{x}}, {\mathbf{u}}):= \begin{bmatrix}
    {\mathbf{x}} -  {\mathbf{x}}_{\max} \\
    -{\mathbf{x}} +  {\mathbf{x}}_{\min} \\
    {\mathbf{u}} -  {\mathbf{u}}_{\max} \\
    -{\mathbf{u}} +  {\mathbf{u}}_{\min}
\end{bmatrix},
\end{equation} 
where $(\mathbf{u}_{\max}, \mathbf{u}_{\min})$ denote the upper and lower bounds on the control inputs, and $(\mathbf{x}_{\max}, \mathbf{x}_{\min})$ define the admissible range of the visual-tactile features, incorporating both the contact-maintenance and perception constraints. \sethlcolor{rev10HL}\hl{The contact is enforced by constraining $\delta$ within the experimentally identified lower and upper bounds, preventing contact loss and excessive sensor deformation against the object.} 

\sethlcolor{rev10HL}\hl{Our VBT-MPC framework considers the following assumptions:}
\begin{itemize}
\item A single dominant contour feature is extracted from the tactile image.
\item The contact is maintained over the prediction horizon.
\item The contacted object is either rigid or semi-rigid, with enough stiffness in its local deformation to generate detectable tactile features. 
\end{itemize}

\subsection{Tactile Contour Extraction and Extended Kalman Filter}
\label{sec:contour_extraction_ekf}

To obtain contour-based tactile features, we have implemented a Neural Network and Line Fitting (NNaLF) approach. It uses a raw RGB tactile image as input. 
This approach implements a first stage based on a segmentation neural network with an encoder-decoder scheme such as Unet++ \cite{Zhou2018}, which has been adapted to obtain contact regions within the input tactile image. Specifically, we use Resnet18 \cite{He2016} as the backbone. Later, a second stage consists in computing the main axis of the segmented region. This is estimated by fitting the segmented region with a straight line using the weighted least-squares algorithm. Finally, the straight line is parametrized to obtain tactile and contour features $(\mathbf{s}_{\pointone}^{\imagesmall},\mathbf{s}_{\pointtwo}^{\imagesmall},\boldsymbol{\xi}^{\imagesmall}
)$, as shown in Section \ref{sec:tactile_features} and illustrated in Fig.~\ref{fig:features_parametrization}.


To estimate contour features $\boldsymbol{\xi}^{\imagesmall}$ at higher rates, we employ an EKF \cite{Lippiello2005}. The tactile points $\mathbf{s}_{\pointone}^{\imagesmall}$ and $\mathbf{s}_{\pointtwo}^{\imagesmall}$ extracted from the VBTS are used as measurements, while the process model is defined by mapping the joint velocities $\dot{\mathbf{q}}$ to the tactile and contour features.

The process model can be defined as follows: 
\begin{equation}
\begin{aligned}
\dot{\hat{\mathbf{x}}}
&=
\begin{bmatrix}
\mathbf{J}_{\pointone}\!\left(\mathbf{s}_{\pointone}^{\imagesmall}\right)\, \mathbf{J}_{r}\left(\mathbf{q}\right) \dot{\mathbf{q}}\\[1mm]
\mathbf{J}_{\pointtwo}\!\left(\mathbf{s}_{\pointtwo}^{\imagesmall}\right)\, \mathbf{J}_{r}\left(\mathbf{q}\right) \dot{\mathbf{q}} \\[1mm]
\mathbf{J}_{\edge}\!\left(\mathbf{s}_{\pointone}^{\imagesmall}, \mathbf{s}_{\pointtwo}^{\imagesmall}\right)\, \mathbf{J}_{r}\left(\mathbf{q}\right) \dot{\mathbf{q}}
\end{bmatrix} +  \mathbf{n},
\end{aligned}
\label{eq:process_model}
\end{equation}
where $(\mathbf{q},\dot{\mathbf{q}}) \in \mathbb{R}^{7}$ are the joint positions and velocities of the robotic manipulator, and $\mathbf{J}_{r}(\mathbf{q}) \in \mathbb{R}^{6 \times 7}$ is the Jacobian manipulator mapping joint velocities to the twist of the end-effector. The term $\mathbf{n} \in \mathbb{R}^{10}$ denotes the noise of the process, modeled as zero-mean white Gaussian noise.

The measurement model is defined as
\begin{equation}
\begin{aligned}
{\mathbf{y}}
&=
\begin{bmatrix}
\mathbf{s}_{\pointone}^{\imagesmall}\\[1mm]
\mathbf{s}_{\pointtwo}^{\imagesmall} \\[1mm]
\end{bmatrix} = \begin{bmatrix}
\mathcal{P}\!\left(\mathbf{x}_{\pointone}^{\camerasmall}, \mathbf{K}\right)\\[1mm]
\mathcal{P}\!\left(\mathbf{x}_{\pointtwo}^{\camerasmall}, \mathbf{K}\right) \\[1mm]
\end{bmatrix} + \mathbf{v},
\end{aligned}
\label{eq:measurement_model}
\end{equation}
where \(\mathbf{v} \in \mathbb{R}^6\) is the noise of the measurement model as zero-mean additive white Gaussian noise.
The EKF is used to propagate the tactile features at the controller frequency of $50~\mathrm{[Hz]}$, even though the GelSight Mini sensor provides measurements at only $15~\mathrm{[Hz]}$. This enables higher-rate and more consistent estimates of the contour features, which can improve the performance of the VBT-MPC controller.


\section{BASELINE METHODS}
\label{sec:base_method}

\subsection{Visual-Tactile Servoing}
By directly applying classical IBVS techniques \cite{Chaumette2016} to tactile sensing and using the contour–feature model \eqref{eq:edge_features}, we define a visual-tactile servoing controller as follows:
\begin{align}
\boldsymbol{\upsilon}^{\camerasmall} & = 
\mathbf{J}^{\dagger}_{\edge}(\boldsymbol{\xi}^{\imagesmall}_e) + (\mathbf{I} - \mathbf{J}^{\dagger}_{\edge} \mathbf{J}_{\edge})\boldsymbol{\lambda},
\label{eq:coupled_controller}
\end{align}
where $\boldsymbol{\lambda}$ \sethlcolor{authorHL}\hl{is the null-space component used to maximize forward velocity along the contour while preserving the minimum-norm solution}.

\subsection{Decoupled Visual-Tactile Servoing}
Because classical IBVS has inherent limitations as presented in Sec.~\ref{Tactile servoing in simulation}, most visual–tactile controllers rely on learned regressions and hybrid force–control schemes~\cite{Aquilina2024, Kappassov2022, LLoyd2024}. \sethlcolor{rev10HL}\hl{These techniques require additional force-control architectures and do not exploit visual servoing directly in the tactile image domain}.
To address this, we implement a decoupled visual-tactile servoing controller 
that extends IBVS to vision-based tactile sensors. We first introduce the decoupled contour feature model in~\eqref{eq:edge_features}, as follows:
\begin{align}
\dot{\boldsymbol{\xi}}^{\imagesmall} &=
\mathbf{J}_{xy}\!\left(\mathbf{s}_{\pointone}^{\imagesmall},\, \mathbf{s}_{\pointtwo}^{\imagesmall}\right)\, \boldsymbol{\upsilon}^{\camerasmall}_{xy}
\;+\;
\mathbf{J}_{z}\!\left(\mathbf{s}_{\pointone}^{\imagesmall},\, \mathbf{s}_{\pointtwo}^{\imagesmall}\right)\, \boldsymbol{\upsilon}^{\camerasmall}_{z},
\label{eq:edge_features_decoupled}
\end{align}
where
$
\boldsymbol{\upsilon}^{\camerasmall}_{xy}
=
\begin{bmatrix}
v_x & v_y & \omega_x & \omega_y
\end{bmatrix}$ and
$\boldsymbol{\upsilon}^{\camerasmall}_{z}
=
\begin{bmatrix}
v_z & \omega_z
\end{bmatrix}$
represent the decoupled components of the end-effector twist. The matrices $\mathbf{J}_{xy} \in \mathbb{R}^{4 \times 4}$ and $\mathbf{J}_{z} \in \mathbb{R}^{4 \times 2}$ are the decoupled interaction matrices corresponding to the columns $\{1,2,4,5\}$ and $\{3,6\}$ of the matrix $\mathbf{J}_{\edge}$, respectively. 
 
Based on the model \eqref{eq:edge_features_decoupled}
, we can define the following control law that enables contour following along the visual tactile features:
\begin{align}
\boldsymbol{\upsilon}^{\camerasmall}_{xy} &= \mathbf{J}^{\dagger}_{xy/r}(r_e -\mathbf{J}_{z/r} ~\boldsymbol{\upsilon}^{\camerasmall}_{z}) + (\mathbf{I} - \mathbf{J}^{\dagger}_{xy/r} \mathbf{J}_{xy/r}) \boldsymbol{\lambda}_{xy} \label{eq:decoupled_control_law_xy},\\
\boldsymbol{\upsilon}^{\camerasmall}_{z} &= \mathbf{J}^{\dagger}_{z/\beta}(\beta_e)  + (\mathbf{I} - \mathbf{J}^{\dagger}_{z/\beta} \mathbf{J}_{z/\beta})
\boldsymbol{\lambda}_{z},
\label{eq:decoupled_control_law_z}
\end{align}
where $\mathbf{J}_{xy/r}$ and $\mathbf{J}_{z/r}$ denote the rows of the matrices $\mathbf{J}_{xy}$ and $\mathbf{J}_{z}$ considering the orthogonal distance $r$ in the contour feature vector. Similarly, $\mathbf{J}_{z/\beta}$ denotes the row of $\mathbf{J}_{z}$ associated with the orientation $\beta$ in the contour feature vector.

The null–space vectors $\boldsymbol{\lambda}_{xy} = \begin{bmatrix}v_{xd} &v_{yd} &\omega_{xd} &
\omega_{yd}\end{bmatrix}^{\top}$ and $\boldsymbol{\lambda}_{z} = \begin{bmatrix}v_{zd} &\omega_{zd}\end{bmatrix}^\top$ represent the desired decoupled components of the twist; where $v_{xd}$ was previously defined in~\eqref{eq:max_vx}. The term $\omega_{yd} = \alpha_d - \alpha$ regulates the tangential orientation of the sensor so that it remains orthogonal to the contact surface, while  $v_{zd} = \delta_d - \delta$ drives the sensor toward the desired local deformation. These null–space vectors allow us to maximize the forward velocity while simultaneously specifying the desired local deformation of the sensor, thereby ensuring continuous contact along the contour–following task. \sethlcolor{authorHL}\hl{To the best of our knowledge, this work presents the first of decoupled visual servoing to vision-based tactile sensors}.

\subsection{Baseline methods for Tactile Contour Extraction}
\label{sec:baseline_extraction}

To compare our contour-based feature extraction method, we implemented three additional approaches based on algorithms from the literature and evaluated both their extraction quality and runtime. All methods take a tactile depth image normalized to the $[0,255]$ range as input, which is first binarized to highlight the contact region. 

\subsubsection{Edge an Line Fitting (EaLF)} The input image is skeletonized and we fit the contour by a line using the weighted least-squares algorithm. \sethlcolor{authorHL}\hl{The position is determined by the midpoint of the line and the orientation is obtained from the collinear vector to the line at the midpoint}.

\subsubsection{Edge and Ellipse Fitting (EaEF)} A Canny edge detector is applied to the input image, followed by a morphological dilation to increase their area and make the contour more representative and connected. Finally, we fit the contour pixels with an ellipse \cite{Andrew1999} then to compute the length and orientation of its major axis and the axis position given by the midpoint between its endpoints.

\subsubsection{Corner and Ellipse Fitting (CaEF)} First, the contact region is obtained by binarization of the input image. Then, most significant corners are extracted using the Shi-Tomasi method \cite{ShiTomasi1994}. Corners that lie too close to each other are discarded and, finally, as in EaEF, an ellipse fitting is applied.

\section{EXPERIMENTS}
\label{sec:experimentation}
\subsection{Setup-up}
\label{sec:setup}
This section presents the experimental validation of our approach. First, we evaluate the tactile contour extraction process by comparing the performance of our NNaLF proposal with the baseline methods described in Section \ref{sec:baseline_extraction}.
Then, we analyze both quality and the runtime of each method to determine its suitability for tactile servo control. Second, we evaluate the proposed \mbox{VBT-MPC} controller in several scenarios and compare it in simulation with the control strategies described in Section \ref{sec:base_method}, such as classic IBVS and decoupled Visual-Tactile servoing. \sethlcolor{authorHL}\hl{In real-world experiments, we compare the proposed method only against the decoupled controller, since classical IBVS fails to maintain constant contact with the object surface}.


\sethlcolor{rev10HL}\hl{
The controllers were implemented in a ROS-based simulation environment, with the same robot manipulator and tactile sensor as used in real-world experiments. This allowed us to test the controllers' responses under kinematic and dynamic conditions similar to those of the real robotic system before transferring them to the physical platform.
}

\sethlcolor{authorHL}\hl{
The robotic system used in the real-world experiments consists of a 7-DoF Kinova Gen3 manipulator equipped with a GelSight Mini VBTS, operating at 15 frames per second with an RGB image of 320 $\times$ 240 $\mathrm{[px]}$ FoV. The sensor uses} a photometric stereoscopic algorithm to reconstruct the depth map of the surface \cite{Yuan2017}. The VBTS was mounted on the end-effector using a custom 3D-printed support, as shown in \sethlcolor{authorHL}\hl{Fig.}~\ref{fig:features_parametrization}. 
\sethlcolor{rev2HL}\hl{
In the real-world experiments, \mbox{VBT-MPC} was tested on several printed-contour types, including a variety of curves (i.e. S-shaped contour) and contours with corners (i.e. hexagon, square). In addition, we tested a wide range of contact conditions on real-world everyday objects with rigid and flexible properties, as well as in multi-contact scenarios.
}


The controllers ran at $50~\mathrm{[Hz]}$, matching the EKF image-processing rate (Section~\ref{sec:contour_extraction_ekf}). The VBT-MPC framework (Fig.~\ref{fig:mpc_pipeline}) was solved by sequential quadratic programming in real time using ACADOS~\cite{acados2022} and CasADi~\cite{2019casadi}, with a $0.5~\mathrm{[s]}$ horizon. \sethlcolor{authorHL}\hl{The optimization constraints that ensure the contact with the object is defined as ($0.0182 \leq \delta \leq 0.022$ m)}. Real-world experiments were conducted in ROS2 on Ubuntu 20.04 using Docker containers.

\subsection{Evaluation of Approaches to Extract Tactile contours}
\begin{figure}[H]
    \centering
     \includegraphics[width=1\columnwidth]{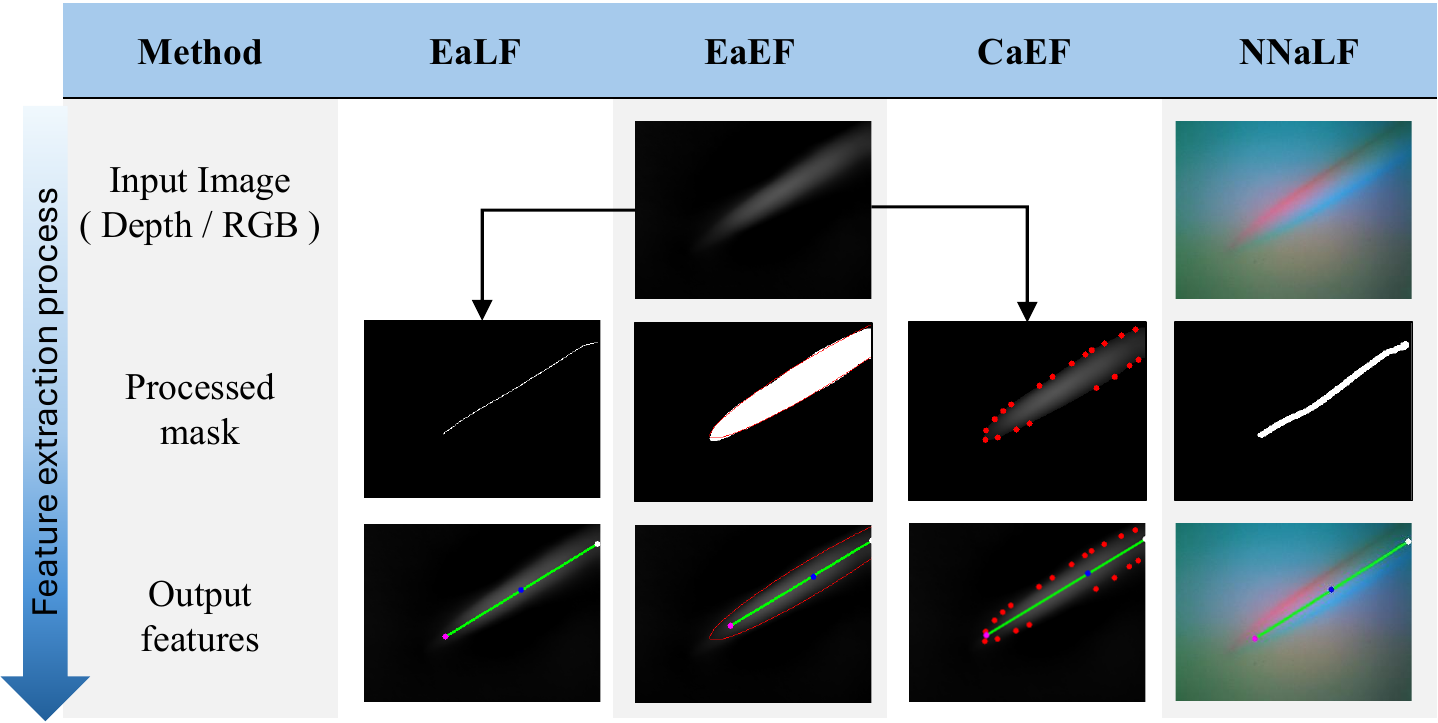}
    \caption{Qualitative results of the approaches evaluated for contour extraction from tactile images.}
    \label{fig:features_extractor_methods}
    \vspace{-15pt}
\end{figure}

In the literature, we did not find datasets of tactile images, in which the images represented contour produced by VBTS. 
\sethlcolor{rev2HL}\hl{Therefore, we generated our own tactile contour dataset for evaluation, which currently contains around 8,000 images acquired with several GelSight Mini sensors. 
The dataset includes contacts on a wide variety of curved and polygonal object edges, as well as challenging multi-contact cases. Reference contour pixels were manually labeled as ground truth to evaluate the contour detection methods.} Fig. \ref{fig:features_extractor_methods} \hl{shows the feature extraction process described} in Sections \ref{sec:contour_extraction_ekf} and \ref{sec:baseline_extraction}.

We used the root mean square error (RMSE) and standard deviation (STD) to evaluate the accuracy 
\sethlcolor{authorHL} \hl {in terms of edge position $\mathrm{[px]}$ and orientation $\mathrm{[rad]}$, both defined in Fig.}~\ref{fig:features_parametrization}. The error is calculated by comparing the contours extracted by each method with the ground truth contour. 
Additionally, the pixel error percentage is reported to quantify the discrepancy between the detected contour and the ground truth line. The results are summarized in Table~\ref{tab:edge_results}.
\begin{table}[H]
\caption{Performance of the approaches implemented to extract features contours with VBTS.}
\resizebox{\columnwidth}{!}{
\begin{tabular}{lccccc|}
\hline
 &\textbf{Position $\mathrm{[px]}$ }&\textbf{Orientation $\mathrm{[rad]}$}&\textbf{Discrepancy}&\textbf{Time $\mathrm{[ms]}$}\\[0.1ex]
 & RMSE $\pm$ STD & RMSE $\pm$ STD & Rate (\%) & RMSE $\pm$ STD \\[0.5ex]
\hline
EaLF  & $22.9 \pm 19.1$ & $0.094 \pm 0.088$ & 19.88 & $76.70 \pm 13.87$ \\[0.5ex]
EaEF  & $25.2 \pm 21.5$ & $0.102 \pm 0.096$ & 27.61 & $75.54 \pm 14.97$ \\[0.5ex]
CaEF  & $24.0 \pm 21.6$ & $0.161 \pm 0.155$ & 14.73 & $57.02 \pm 1.32$ \\[0.5ex]
NNaLF & $\mathbf{10.4 \pm 7.9}$ & $\mathbf{0.034 \pm 0.025}$ & \textbf{9.15} & $\mathbf{28.84 \pm 7.53}$ \\[0.5ex]
\hline
\end{tabular}
}
\label{tab:edge_results}

\end{table}
\begin{figure}[H]
    \centering
    \includegraphics[width=1\linewidth]{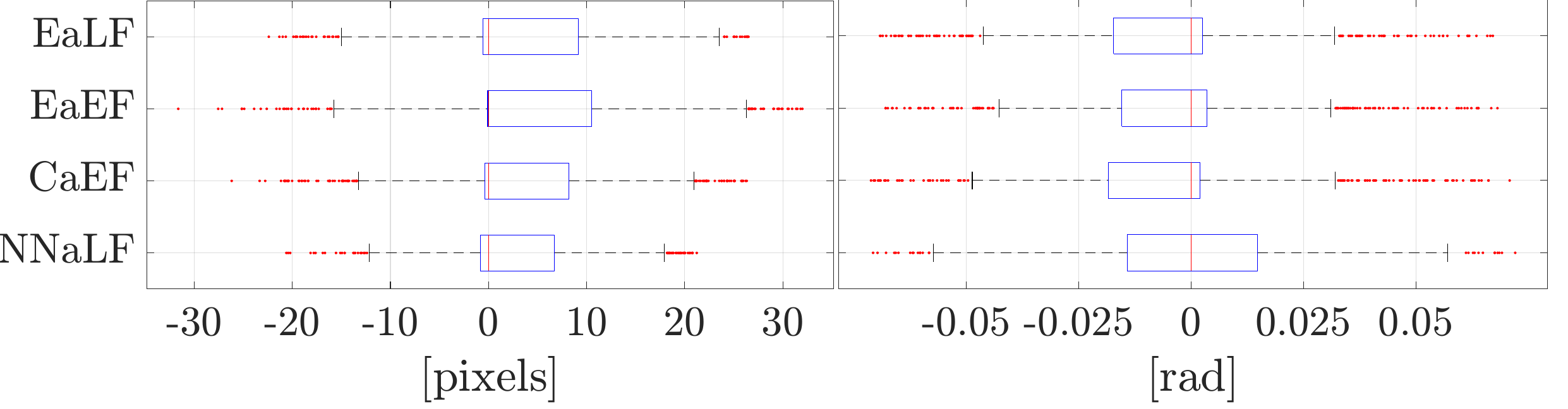}
    \caption{Statistical study of position (left) and orientation error (right) for each approach shown in Table \ref{tab:edge_results}.} 
    \label{fig:box_plot}

\end{figure}
\sethlcolor{authorHL}\hl{As shown in Fig.}~\ref{fig:box_plot}\hl{, NNaLF exhibits the lowest dispersion in the position error, with a narrower interquartile range, indicating lower variability and more consistent performance. For the orientation errors, NNaLF shows greater variability; however, it still attains the lowest value of this error, as reported in Table}~\ref{tab:edge_results}\hl{, making this dispersion acceptable. Moreover, its error distribution is more symmetric around the median, while the other methods exhibit more outliers.}  \sethlcolor{rev2HL}\hl{It is also important to note that the non-learning-based methods cannot properly handle multi-contact cases, while NNaLF is able to detect the target contour under these conditions. Overall, these results support NNaLF as the most suitable method for this task due to 
its ability to operate in multi-contact scenarios and with a variety of types of contours.}




\subsection{Tactile servoing in simulation}
\label{Tactile servoing in simulation}
In this section, we compare the proposed \mbox{VBT-MPC} controller to other baseline controllers, using synthetic tactile features generated by touch.

The evaluation focuses on the controller ability to maintain constant touch with the surface by driving the tactile features toward their desired values while simultaneously achieving the desired forward velocity. For this test, \sethlcolor{authorHL}\hl{the desired forward velocity is defined as} $v_{xd} = 0.5~\mathrm{[cm/s]}$, and the contour to be followed was defined as a straight edge without curvature. 
\begin{figure}[b]
     \centering
    \includegraphics[width=0.95\linewidth]{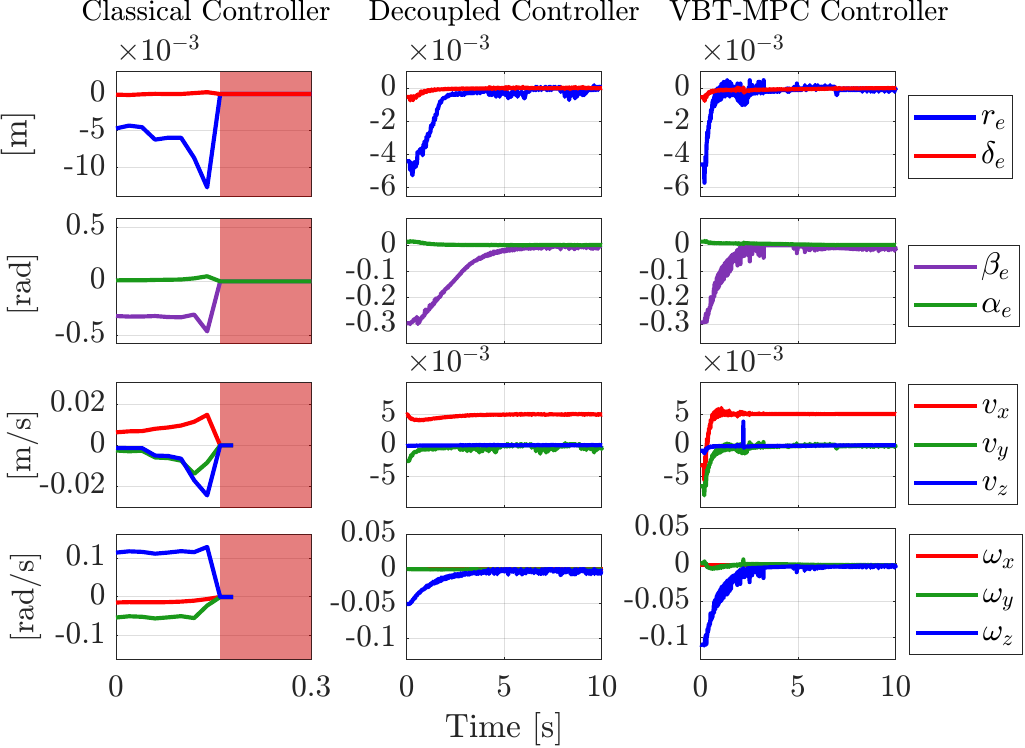}
     \caption{\sethlcolor{rev10HL}\hl{Comparison of tactile feature errors and control velocities under simulation experiments.}}
      \label{fig:simulation_Experiment} 

\end{figure}
The contour errors and velocities obtained from in the simulation experiments are shown in Fig. \ref{fig:simulation_Experiment}. The results show that when the classic tactile visual servoing controller attempts to reduce the feature errors, the control actions quickly move the tactile sensor away from the surface (red zone in Fig. \ref{fig:simulation_Experiment}). This fact causes a loss of contact, which prevents it from following the contour.
\sethlcolor{rev10HL}\hl{This is a well known limitation of IBVS (camera retreat), which can induce large motions and drive the sensor away from the target along the normal} \cite{corke2002}.
In contrast, both the decoupled controller and our \mbox{VBT-MPC} strategy converge to the desired feature values and reach the desired forward velocity $v_{xd}$. The decoupled controller achieves this by separating the forward motion from the characteristic regulation, thus avoiding the divergence observed in the classical approach, but it is not possible to apply restrictions to image features and control inputs.

 \sethlcolor{authorHL}\hl{Compared with the decoupled controller, \mbox{VBT-MPC} reduces the settling time by about $1.0 ~\mathrm{[s]}$ while reaching the desired forward velocity $v_{xd}$}. 
This shows that our proposal offers better performance for the task of following a straight edge without curvature.

\begin{figure*}[!ht]
    \centering
    \begin{minipage}[c]{0.22\linewidth}
        \centering
        \subfloat[]{\label{fig:kinova_s_shape}
            \includegraphics[width=\linewidth]{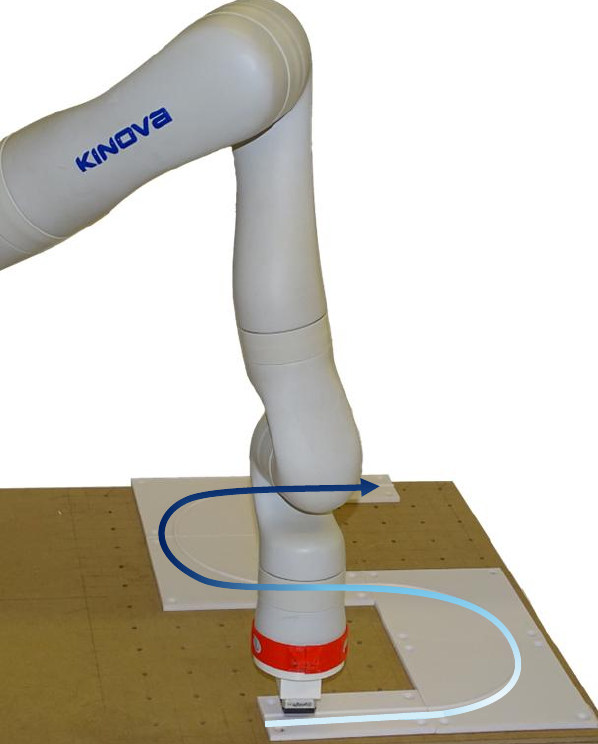}}
    \end{minipage}
    \hfill
    \begin{minipage}[c]{0.35\linewidth}
        \centering
        \subfloat[]{\label{fig:path_comparation}
            \includegraphics[width=\linewidth]{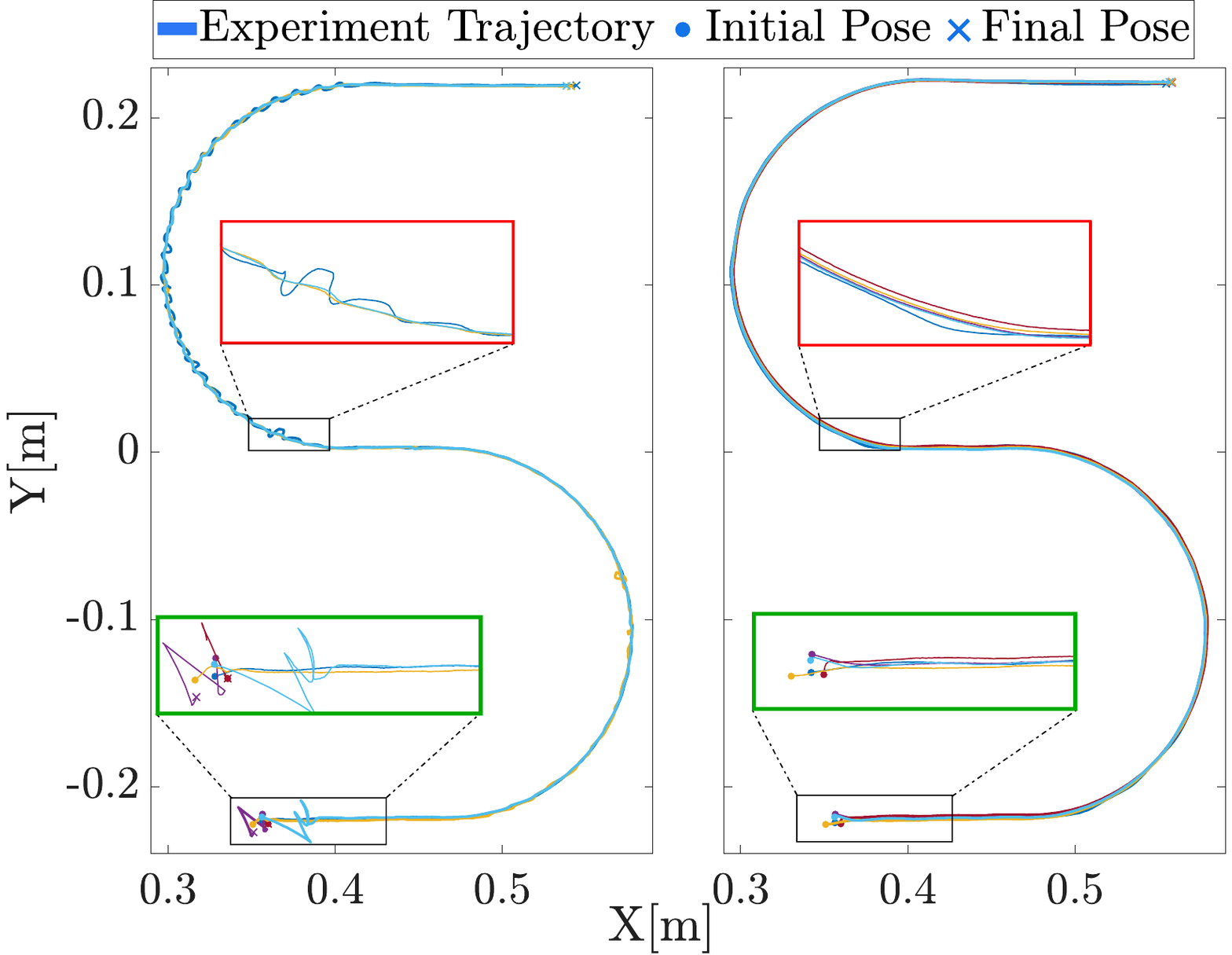}}
    \end{minipage}
    \hfill
    \begin{minipage}[c]{0.17\linewidth}
        \centering
        \subfloat[]{\label{fig:errors_comparation}
            \includegraphics[width=\linewidth]{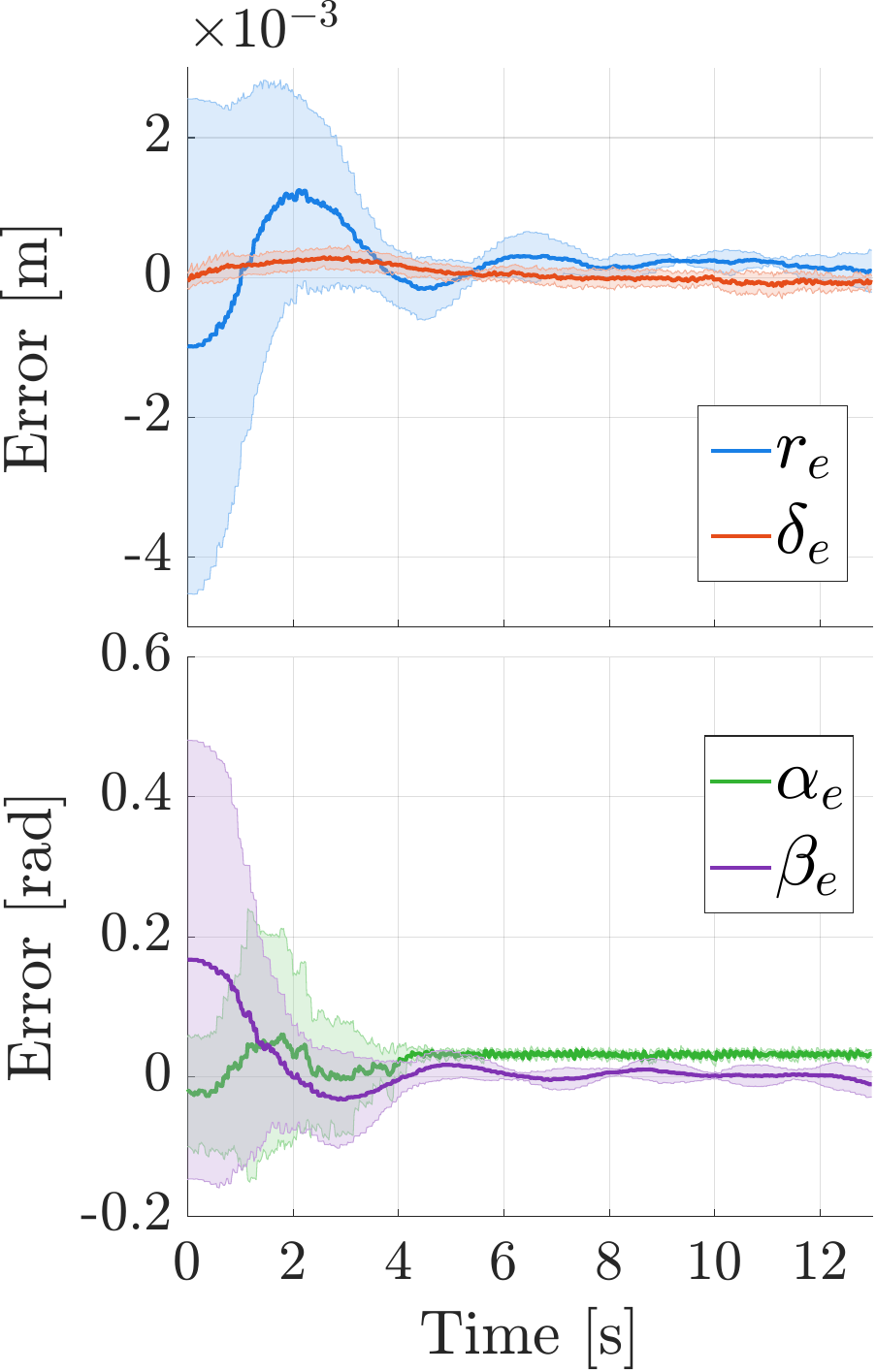}}
    \end{minipage}
    \hfill
    \begin{minipage}[c]{0.16\linewidth}
        \centering
        \subfloat[]{\label{fig:velocities_comparation}
            \includegraphics[width=\linewidth]{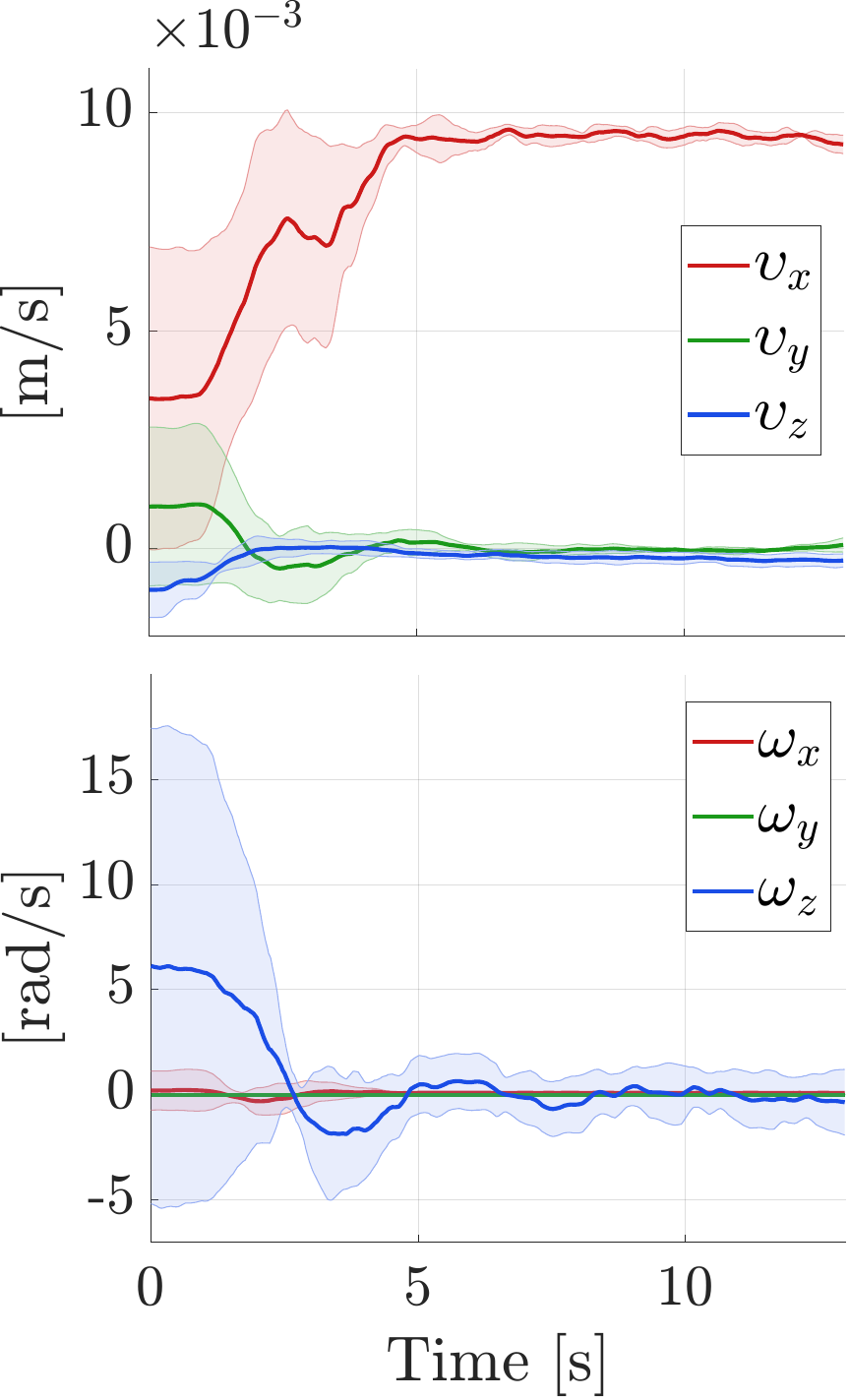}}
    \end{minipage}  
    \vspace{-5pt}
    \caption{Comparison between the decoupled controller and VBT-MPC during real-robot tracking of a 3D-printed S-shaped contour. \sethlcolor{authorHL}\hl{\textbf{(a)} Kinova Gen3 with GelSight sensor following the contour.} \textbf{(b)} Trajectories over five trials for each controller (left: decoupled controller, right: VBT-MPC). Green boxes indicate the initial straight segment and red boxes the response to an external perturbation. \textbf{(c)} Tactile features errors, and \textbf{(d)} linear and angular velocities of VBT-MPC, respectively.}
    \label{fig:3d_path_comparation}
\end{figure*}

\subsection{Tactile servoing in real-world}
\subsubsection{Experiments on 3D-Printed Contours}

Real-world experimentation is focused on the decoupled controller and the proposed \mbox{VBT-MPC} method, as the classical controller demonstrated an inability to maintain physical contact in simulation. Both controllers are evaluated for S-shaped contour following, which was printed on a $XY$ flat surface (see Fig.\ref{fig:kinova_s_shape}). A total of five trials are conducted for each controller. In order to evaluate our proposal under varying initial conditions, we use a different starting position. The five trajectories carried out by the robot, using each of two controllers, are shown in Fig~\ref{fig:path_comparation}. The results show that the robot can follow the shape using either controller. However, compared to \mbox{VBT-MPC}, the decoupled approach manifests noticeable oscillations and deviations during the initial straight sections of the path as observed in the green boxes of Fig~\ref{fig:path_comparation}. In particular, the decoupled controller exhibits significant variability between trials. In two experiments, the trajectory was not completed successfully due to oscillations. They caused a loss of contact between the sensor and the object surface. In contrast, the VBT-MPC generates similar trajectories (quasi-overlapping lines) and, in all cases, the robot successfully followed the 3D-printed contour.


In addition, the errors in the tactile features and control velocities for the proposed VBT-MPC strategy are shown in Figs.\ref{fig:errors_comparation} and \ref{fig:velocities_comparation}. The graphs illustrate both the mean and the standard deviation for the first $13~\mathrm{[s]}$ of the five tests. As can be seen, the tactile errors converge towards a value close to zero in all tests performed. In quantitative terms, RMSE remains below $4.8\times10^{-4}~\mathrm{[m]}$ for $r_e$ and $2.0\times10^{-4}~\mathrm{[m]}$ for $\delta_e$, while the angular RMSE are $3.0\times10^{-2}~\mathrm{[rad]}$ for $\alpha_e$, and $4.6\times10^{-2}~\mathrm{[rad]}$ for $\beta_e$. These values confirm sub-millimeter contour tracking with small orientation errors, in agreement with the trends observed in Fig~\ref{fig:errors_comparation}. This result proves the controller's ability to compensate the variations in the initial posture. 
\sethlcolor{authorHL}\hl{Furthermore, the forward velocity component $v_x$ closely matches the desired value $v_{xd}=1.0~\mathrm{[cm/s]}$, with low dispersion across experimental trials. The remaining velocity components stay close to zero, as expected from the constrained optimization process, as shown in Fig.}~\ref{fig:velocities_comparation}.

During the tests, external disturbances were deliberately introduced to assess the behavior of the controllers. The red box in Fig. \ref{fig:path_comparation} 
\hl{shows the responses of both controllers to disturbances applied during the third trial}. The decoupled controller attempted to return to the nominal trajectory; however,  this resulted in larger oscillations that persisted throughout the curved segment. 
\hl{Although the trajectory was eventually recovered along the final straight section of the 3D-printed reference contour, the transient response remained oscillatory}. In contrast, the VBT-MPC shows only a minor transient deviation from the reference trajectory and is successfully realigned with the intended trajectory.

\begin{figure}[t]
    \centering
    \begin{minipage}[c]{0.47\linewidth}
        \centering
       \makebox[\linewidth][r]{%
        \includegraphics[width=0.84\linewidth]{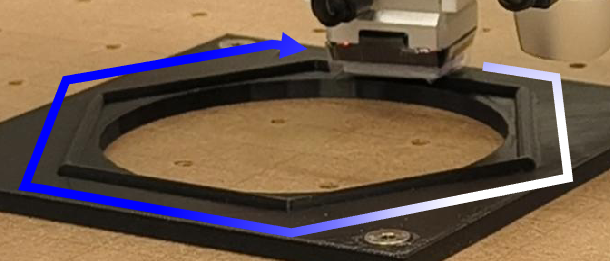}} 
        
        \vspace{1pt}   
        
        \includegraphics[width=\linewidth]{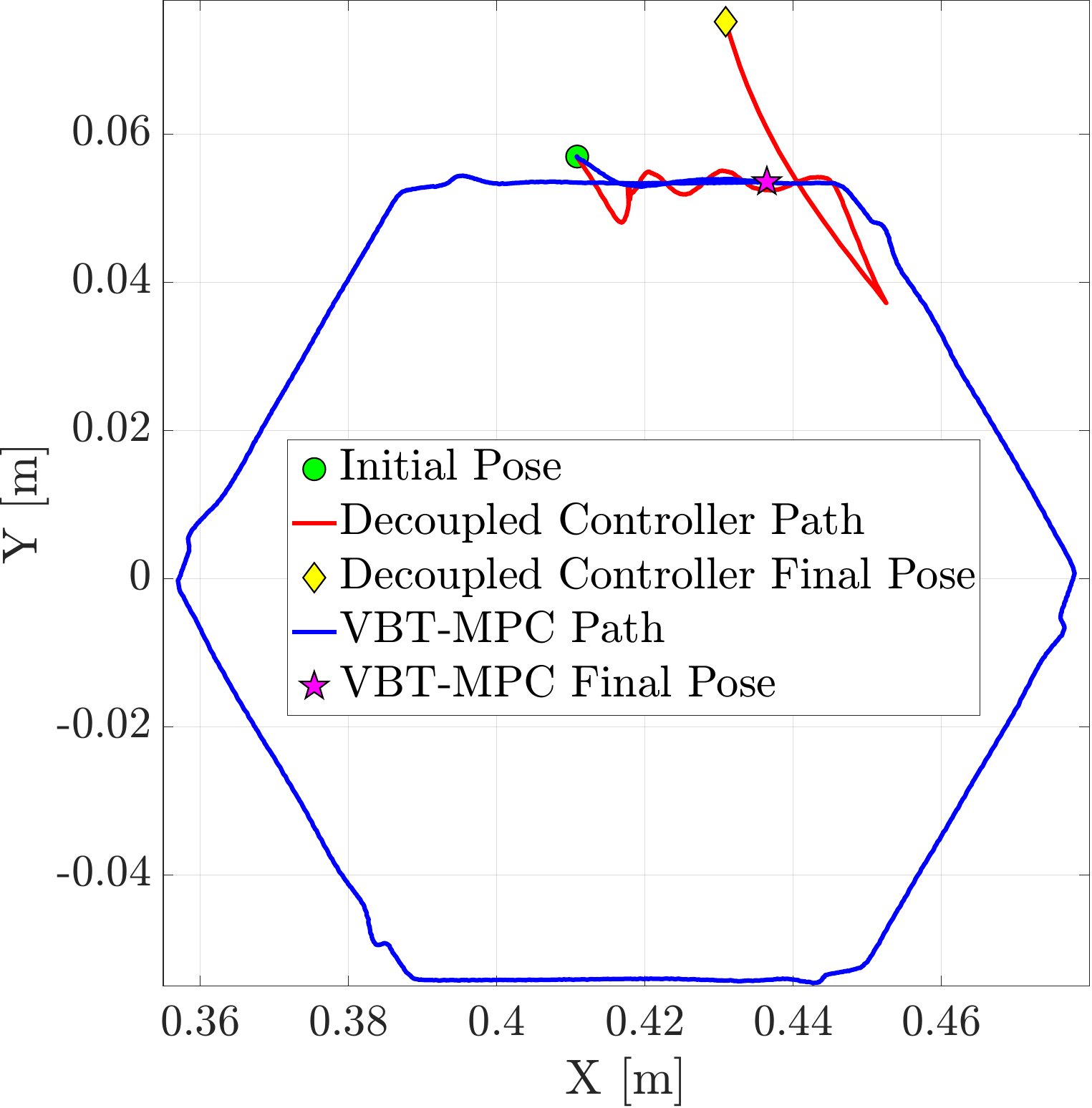}
    \end{minipage}
    \hfill
    \begin{minipage}[c]{0.45\linewidth}
        \centering
        \includegraphics[width=\linewidth]{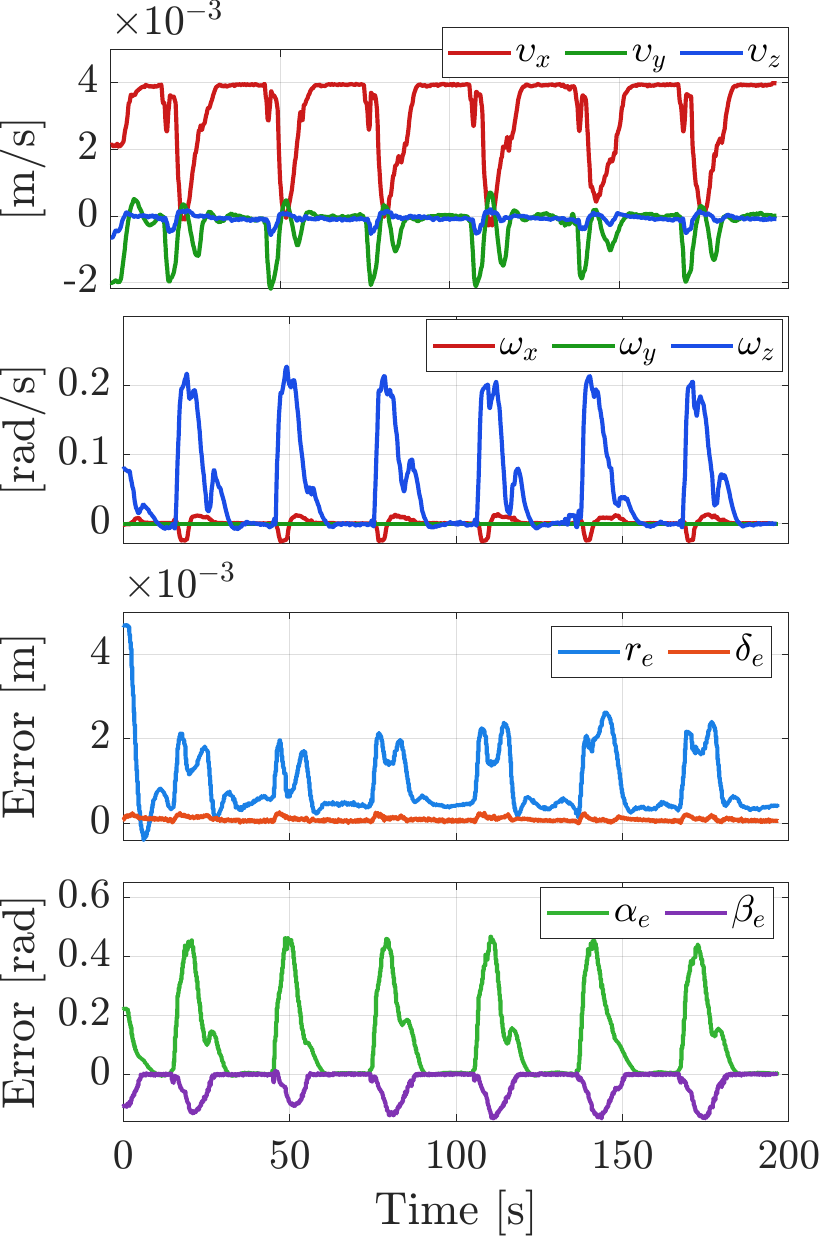}
    \end{minipage}
    \caption{\sethlcolor{rev2HL}\hl{Hexagonal contour-following comparison. Paths for both controllers, and VBT-MPC velocities and errors.}}
    \label{fig:hex_exp}
\end{figure}

\sethlcolor{rev2HL}\hl{
In addition, both controllers are tested on a 3D-printed hexagonal contour with raised edges and corners (see Fig.}~\ref{fig:hex_exp}), \hl{at a desired forward velocity of $v_{xd} = 0.4~\mathrm{[cm/s]}$, to evaluate tracking during abrupt corner transitions.

As shown in Fig} \ref{fig:hex_exp}\hl{, the VBT-MPC successfully completes the hexagonal trajectory, maintaining consistent contour tracking along the entire edge, including corner transitions.  In contrast, the decoupled controller fails at the first corner and it is unable to continue tracking the edge. This demonstrates the advantages of including constraints that enforce the tactile features to remain within the FoV of the VBTS.}


\begin{figure*}
    \centering
    \begin{subfigure}{0.19\linewidth}
        \centering
        \includegraphics[width=\linewidth]{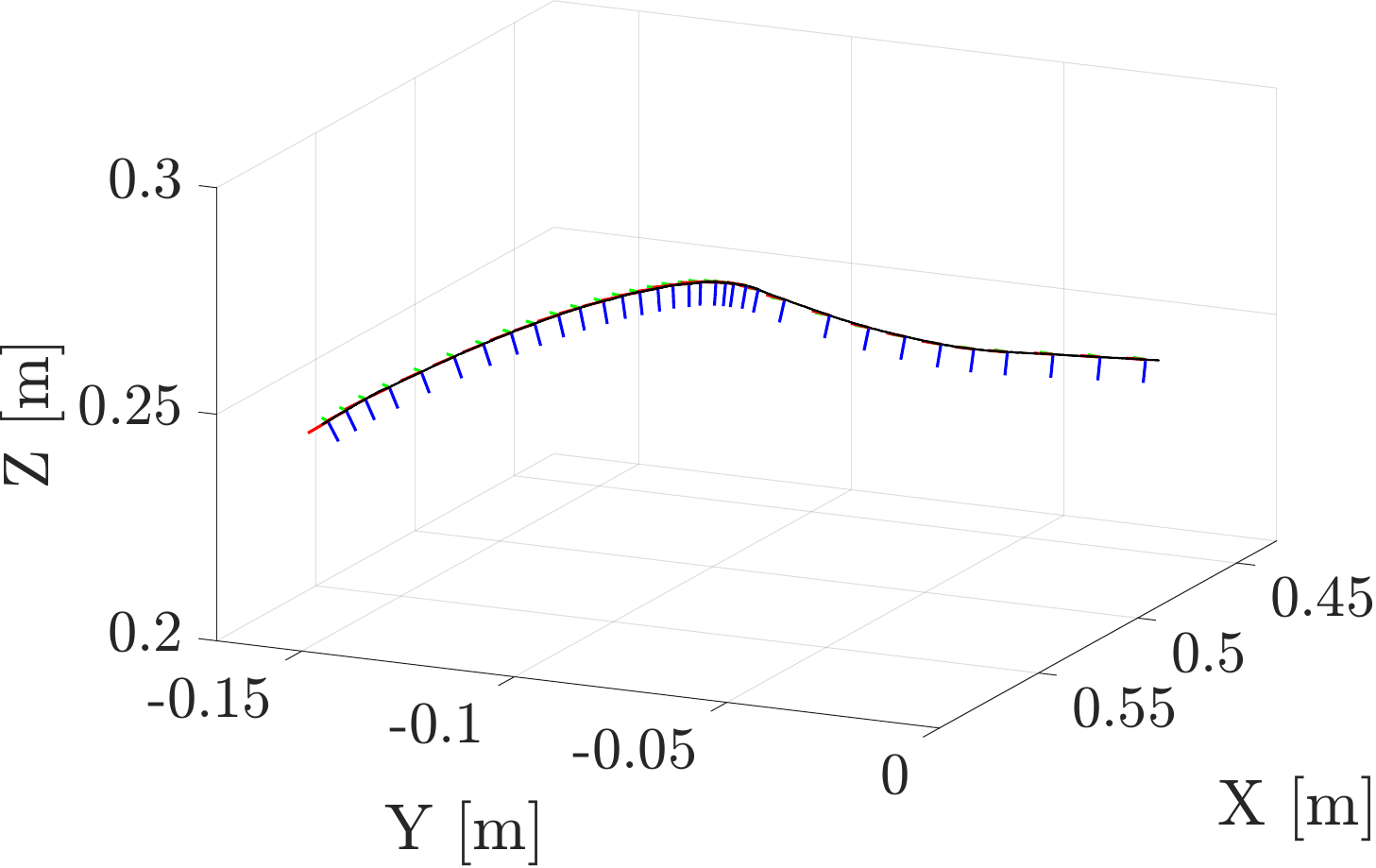}
        \label{fig:horma_path}
    \end{subfigure}
    \hfill
        \begin{subfigure}{0.19\linewidth}
        \centering
        \includegraphics[width=\linewidth]{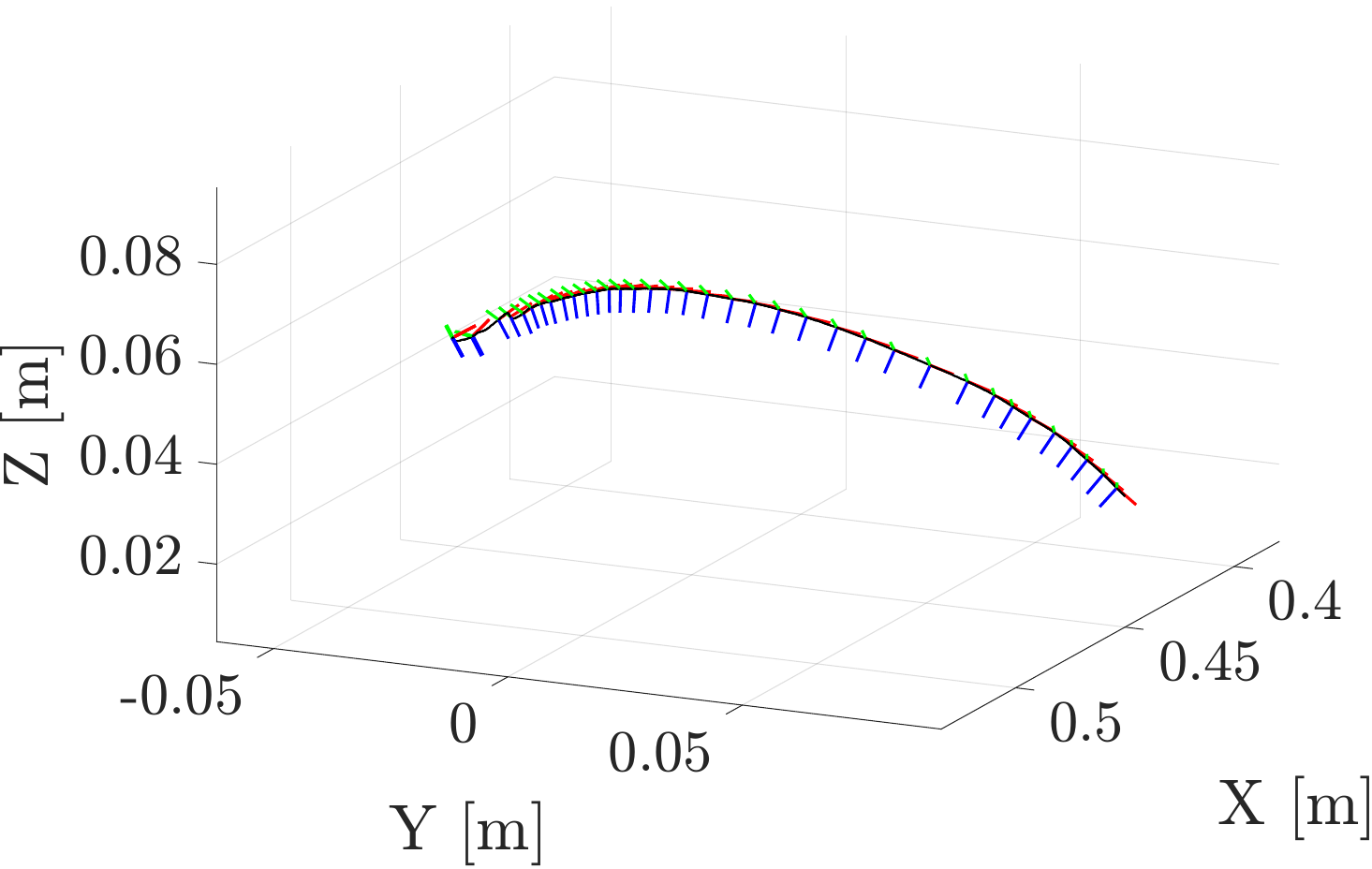}
        \label{fig:banana_path}
    \end{subfigure}
    \hfill
        \begin{subfigure}{0.19\linewidth}
        \centering
        \includegraphics[width=\linewidth]{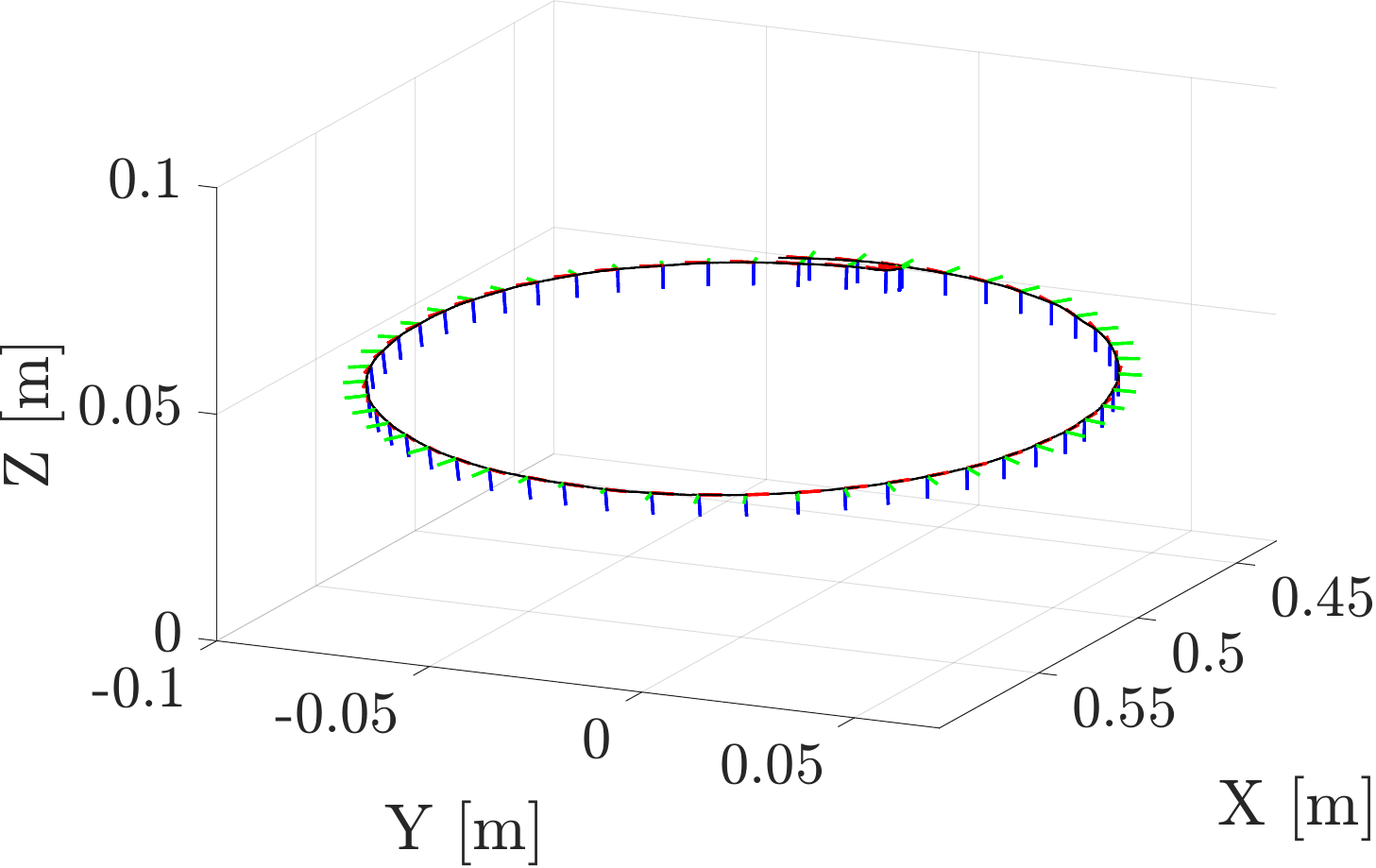}
        \label{fig:plate_path}
    \end{subfigure}
    \hfill
        \begin{subfigure}{0.19\linewidth}
        \centering
        \includegraphics[width=\linewidth]{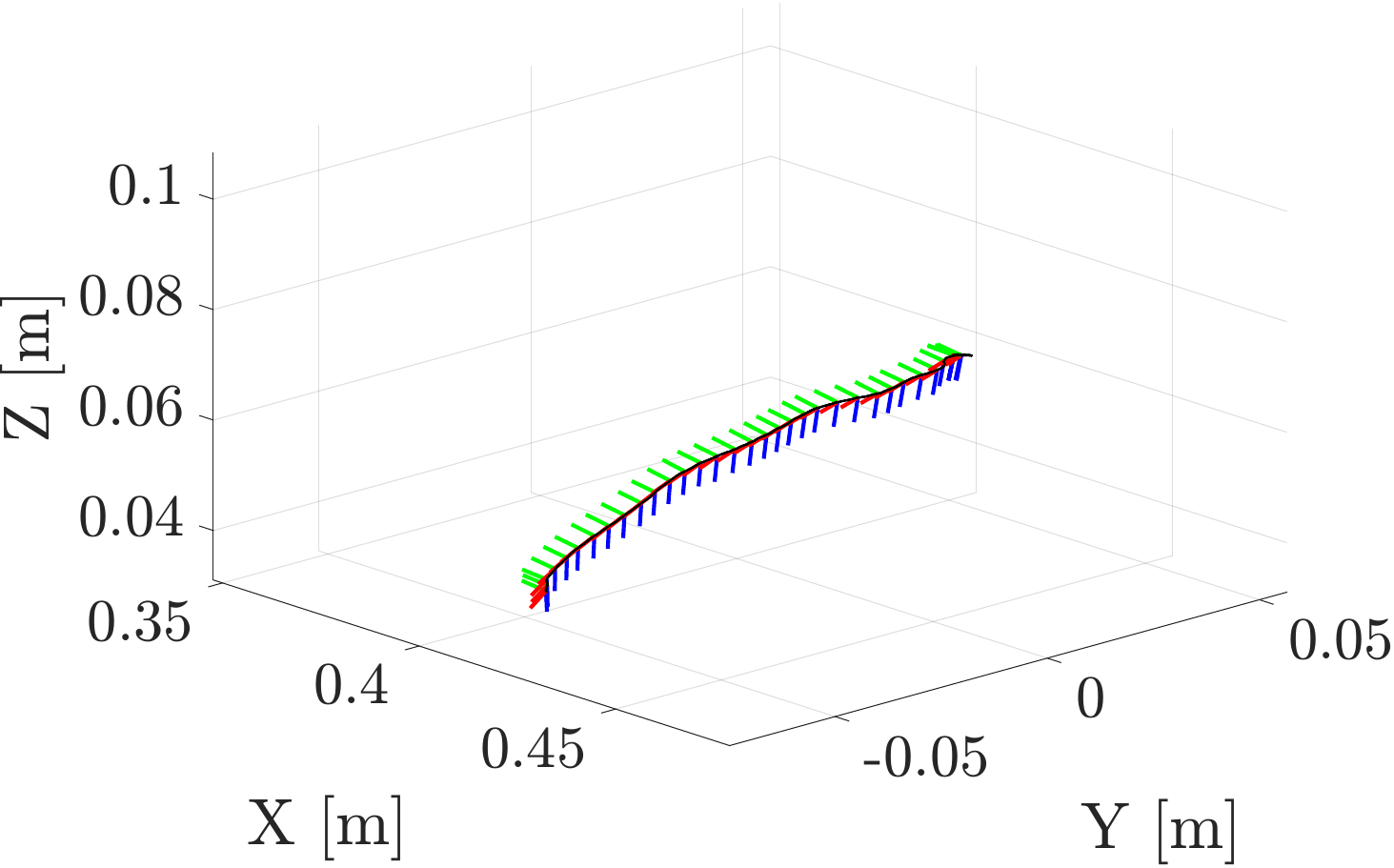}
        \label{fig:zipper_path}
        \end{subfigure}%
    \hfill
    \begin{subfigure}[b]{0.19\linewidth}
        \centering
        \includegraphics[width=\linewidth]{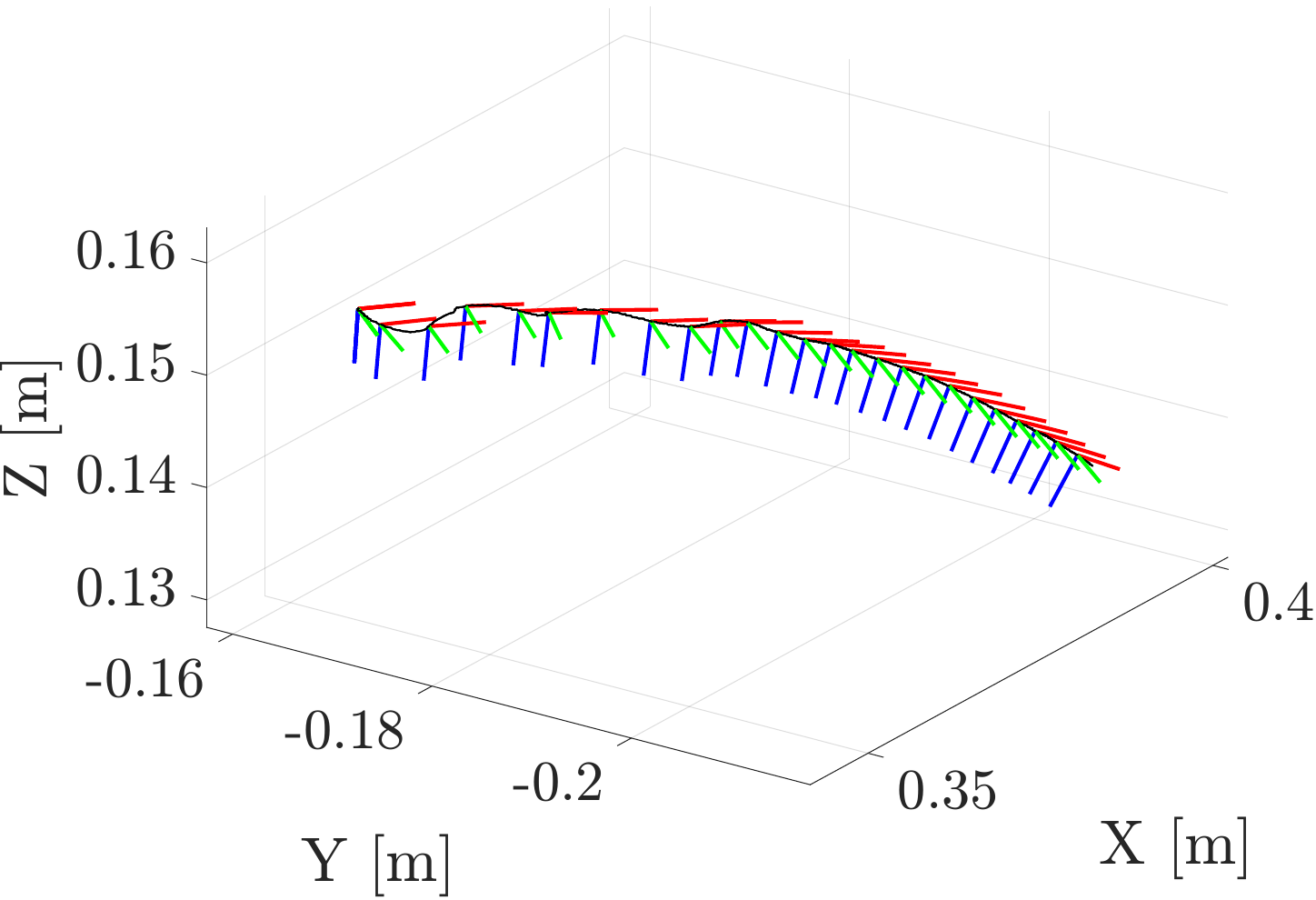}
        \label{fig:seam_path}
    \end{subfigure}\\[-8pt]
    \begin{subfigure}[b]{0.19\linewidth}
        \centering
        \includegraphics[width=\linewidth]{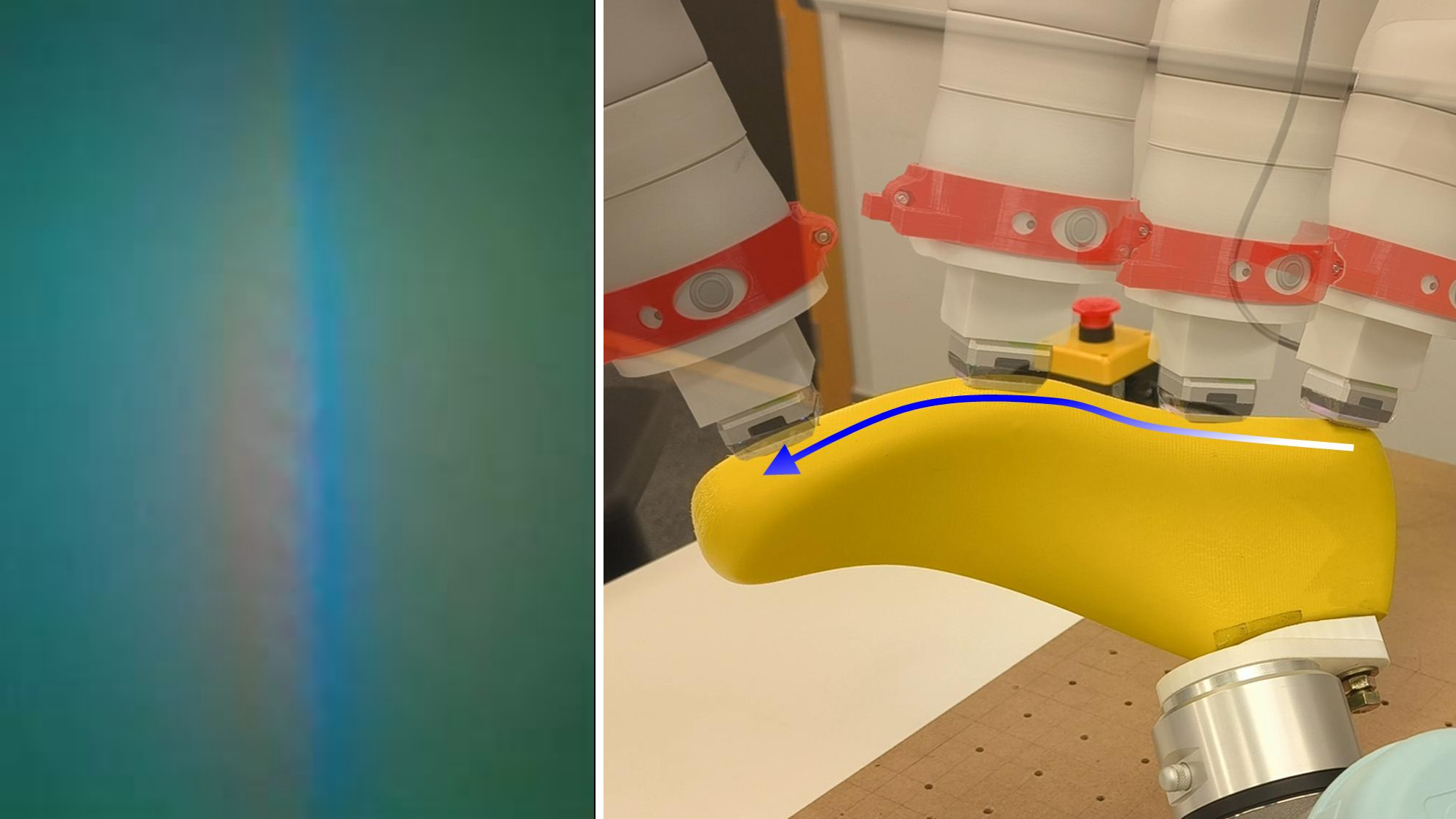}
        \label{fig:shoe_last_and_gs}
    \end{subfigure}
    \hfill
    \begin{subfigure}{0.19\linewidth}
        \centering
        \includegraphics[width=\linewidth]{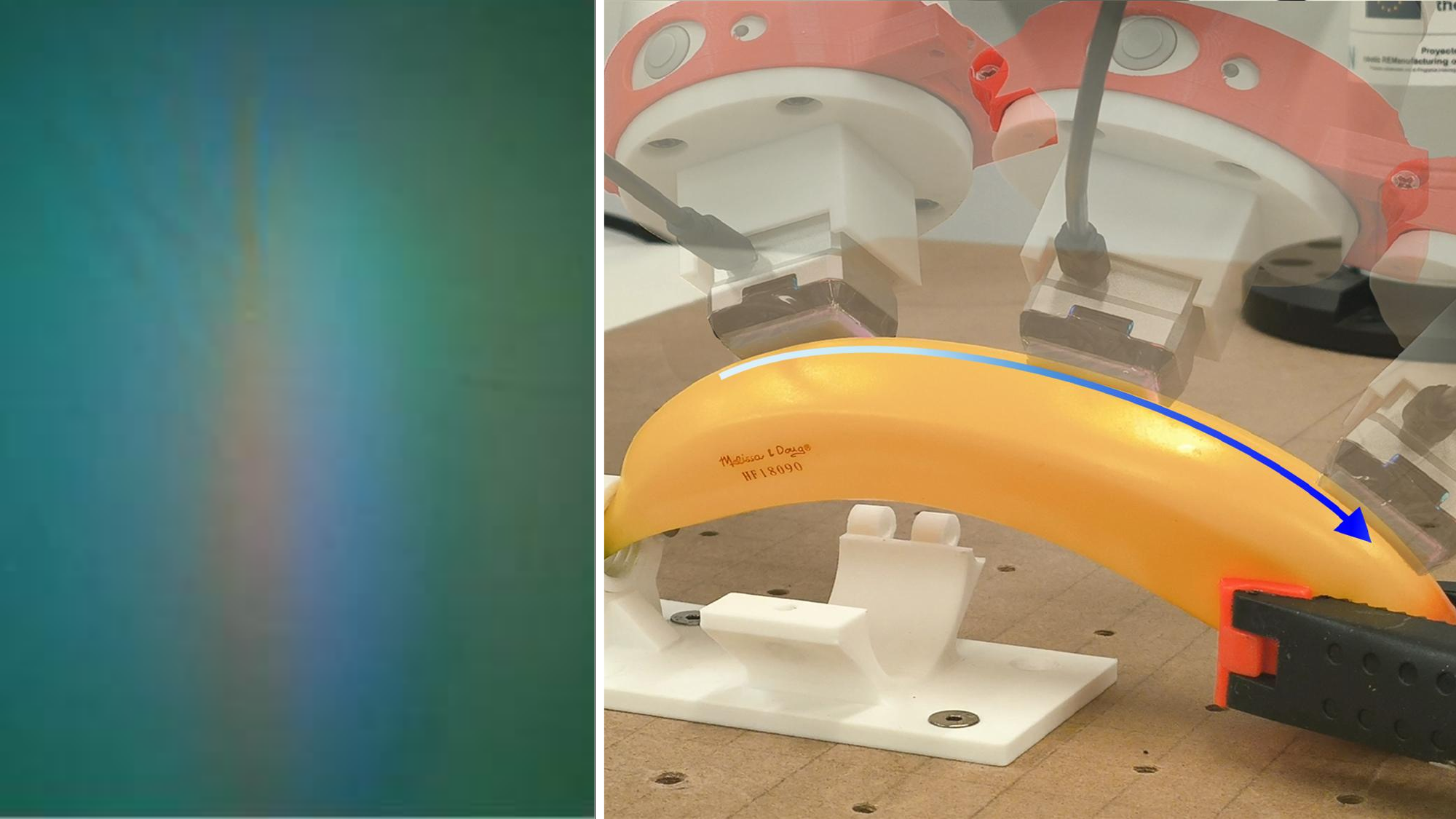}
        \label{fig:banana_and_gs}
    \end{subfigure}
    \hfill
    \begin{subfigure}{0.19\linewidth}
        \centering
        \includegraphics[width=\linewidth]{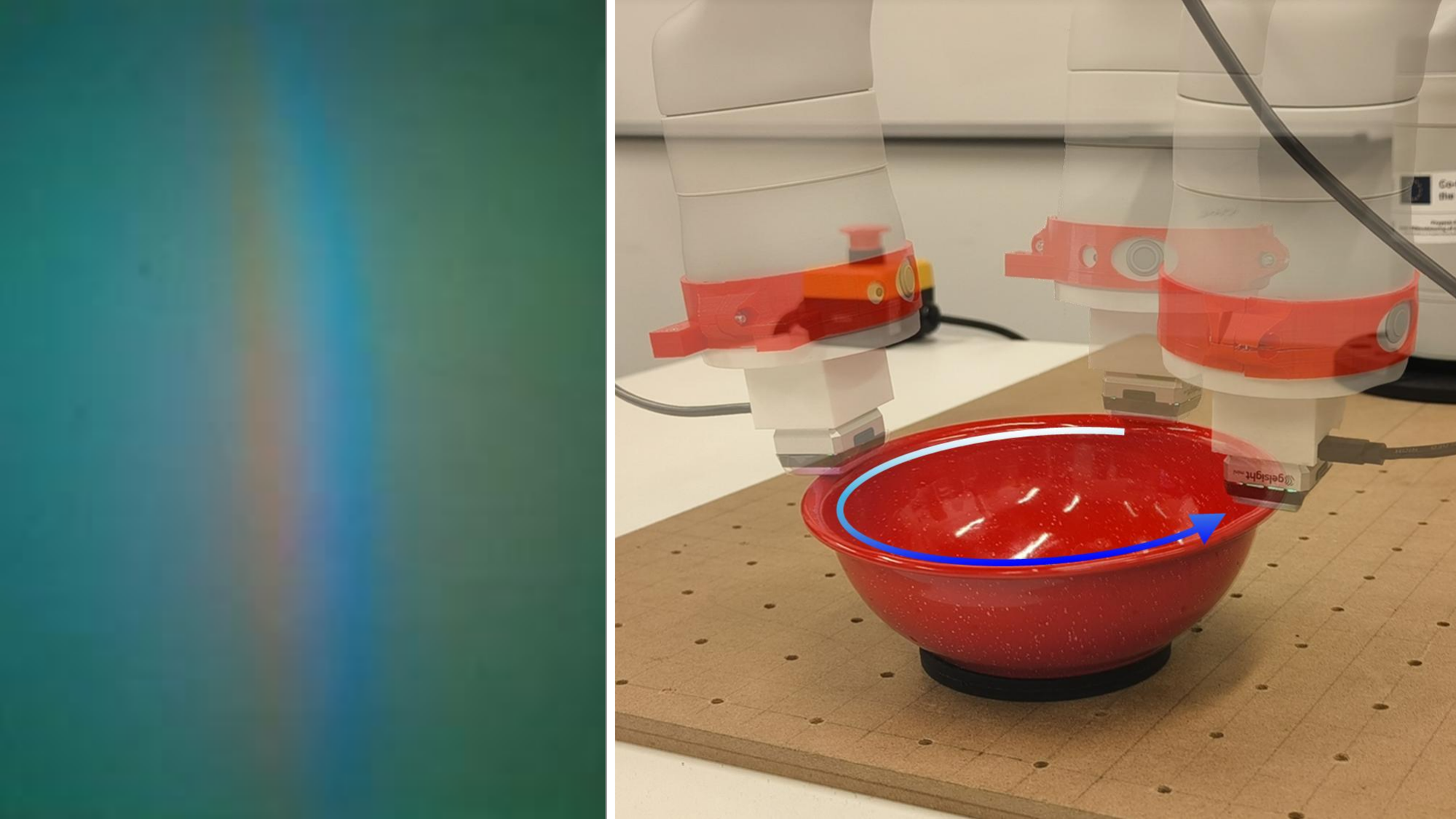}
        \label{fig:plate_and_gs}
    \end{subfigure}
    \hfill
    \begin{subfigure}{0.19\linewidth}
        \centering
        \includegraphics[width=\linewidth]{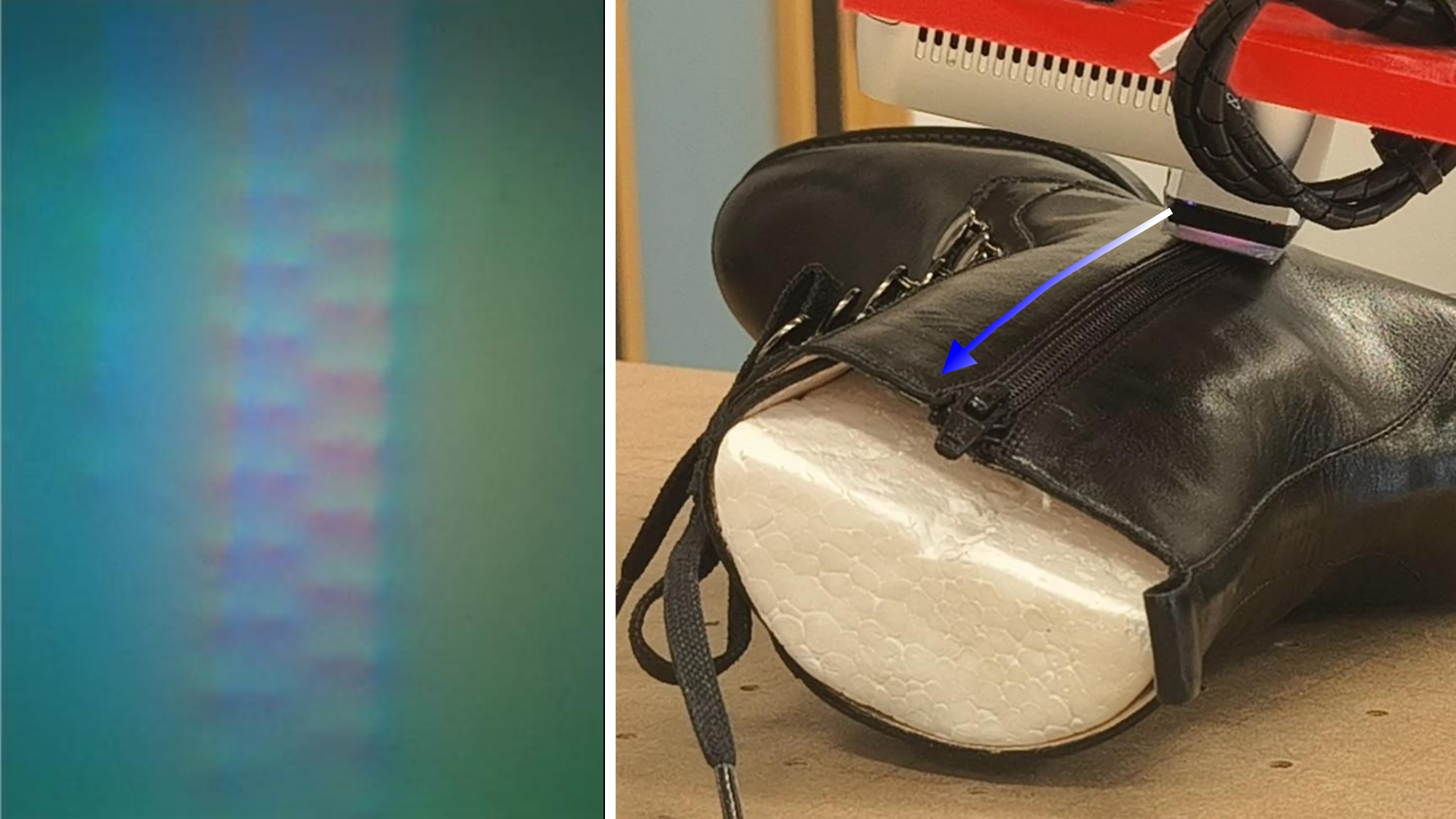}
        \label{fig:zipper_and_gs}
    \end{subfigure}   
        \hfill
    \begin{subfigure}{0.19\linewidth}
        \centering
        \includegraphics[width=\linewidth]{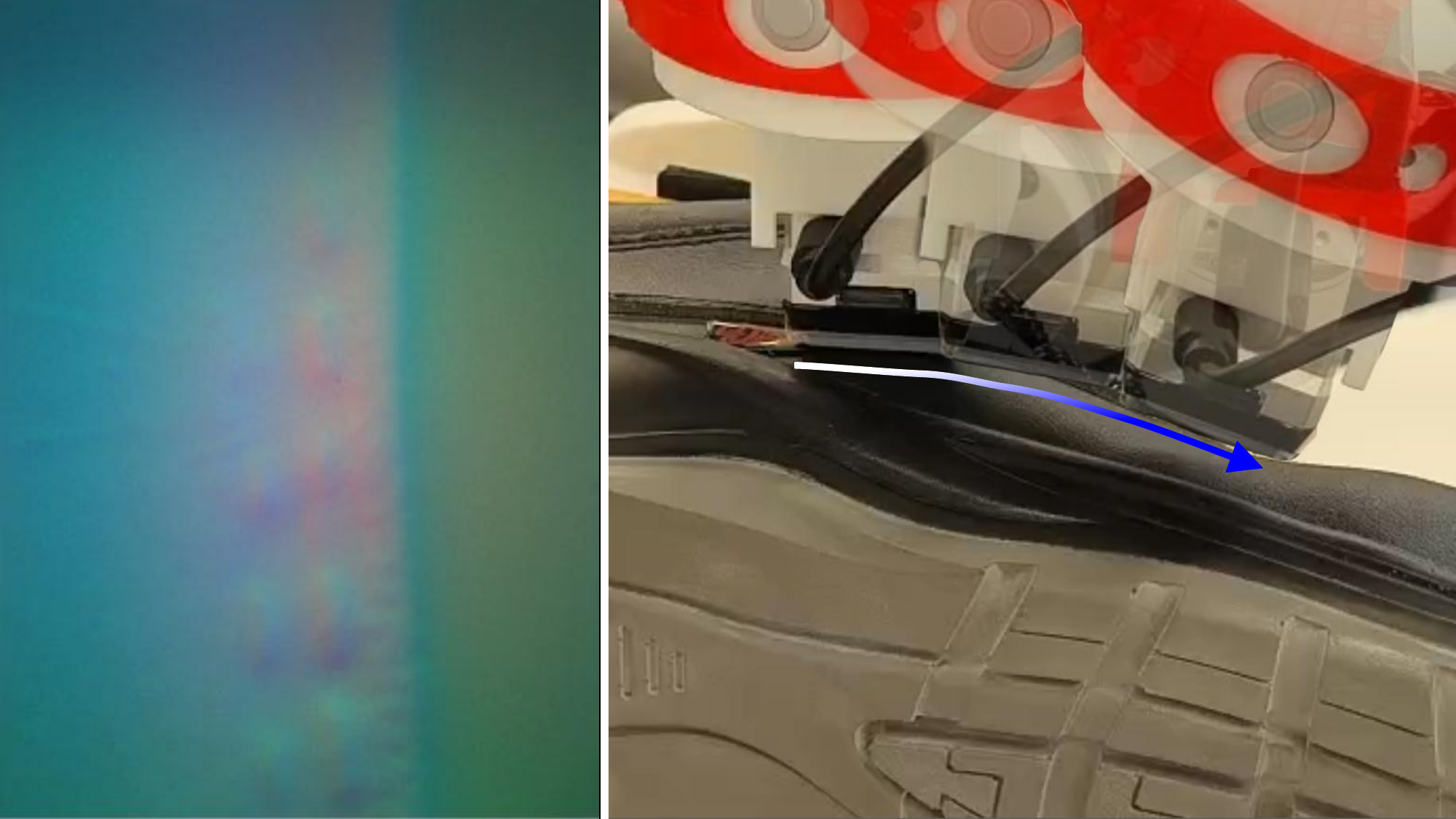}        
        \label{fig:seam_and_gs}
    \end{subfigure}

    \caption{\sethlcolor{authorHL}\hl{Tactile contour-following experiments using the \mbox{VBT-MPC} method. (up) Show reconstructed trajectories; (bottom) display the corresponding objects and a sample of what the tactile sensor records. Arrows indicate the motion direction.}} \label{fig:real_objects}
\end{figure*}
\begin{table*}
    \centering
    \caption{RMSE of contour-following errors and mean computer computational time for real-world objects.}
    \label{tab:real_objects_rmse}
    \resizebox{\linewidth}{!}{
    \begin{tabular}{lccccc}
        \hline
        Object & $r_e~\mathrm{[m]}$ & $\delta_e~\mathrm{[m]}$ & $\alpha_e~\mathrm{[rad]}$ & $\beta_e~\mathrm{[rad]}$ & \sethlcolor{rev10HL}\hl{Solving time$~\mathrm{[ms]}$}\\
        \hline     
        Shoe last 
            & $1.54\times10^{-3} \pm 1.52\times10^{-3}$ 
            & $7.01\times10^{-5} \pm 7.00\times10^{-5}$ 
            & $2.73\times10^{-2} \pm 2.70\times10^{-2}$ 
            & $3.48\times10^{-2} \pm 2.78\times10^{-2}$
            & \sethlcolor{rev10HL}\hl{$20.59 \pm 0.79$}\\
         Plastic banana 
            & $2.02\times10^{-3} \pm 2.02\times10^{-3}$ 
            & $1.55\times10^{-3} \pm 1.55\times10^{-3}$ 
            & $2.73\times10^{-2} \pm 2.03\times10^{-2}$ 
            & $1.06\times10^{-1} \pm 9.58\times10^{-2}$
            & \sethlcolor{rev10HL}\hl{$20.64 \pm 0.81$}\\
        Plate 
            & $3.98\times10^{-4} \pm 3.35\times10^{-4}$ 
            & $7.45\times10^{-5} \pm 4.88\times10^{-5}$ 
            & $2.73\times10^{-2} \pm 1.33\times10^{-2}$ 
            & $6.12\times10^{-2} \pm 4.71\times10^{-2}$
            & \sethlcolor{rev10HL}\hl{$20.54 \pm 0.78$}\\
        Boot zipper 
            & \sethlcolor{rev2HL}\hl{$5.11\times10^{-4} \pm 4.96\times10^{-4}$} 
            & \sethlcolor{rev2HL}\hl{$4.58\times10^{-5} \pm 4.09\times10^{-5}$}
            & \sethlcolor{rev2HL}\hl{$2.68\times10^{-2} \pm 2.12\times10^{-2}$}
            & \sethlcolor{rev2HL}\hl{$5.81\times10^{-2} \pm 5.28\times10^{-2}$}
            & \sethlcolor{rev10HL}\hl{$20.53 \pm 0.96$}\\
        Shoe seam 
            & \sethlcolor{rev2HL}\hl{$1.20\times10^{-3} \pm 1.06\times10^{-3}$}
            & \sethlcolor{rev2HL}\hl{$1.26\times10^{-4} \pm 7.55\times10^{-5}$}
            & \sethlcolor{rev2HL}\hl{$2.59\times10^{-2} \pm 2.51\times10^{-2}$}
            & \sethlcolor{rev2HL}\hl{$2.80\times10^{-2} \pm 2.75\times10^{-2}$}
            & \sethlcolor{rev10HL}\hl{$20.74 \pm 0.92$}\\
        \hline        
    \end{tabular}    
    }
\end{table*}
\subsubsection{Experiments on real-world Objects}
Additionally, VBT-MPC was evaluated on \sethlcolor{rev2HL}\hl{several} real-world objects with different geometries \sethlcolor{rev2HL}\hl{and stiffness properties} to demonstrate its ability to generalize beyond contours on flat surfaces. In each experiment, the robot used visual tactile feedback to trace the contour of the object at a constant forward velocity \mbox{$v_{x,\max} = 0.5~\mathrm{[cm/s]}$} \sethlcolor{authorHL}\hl{for the shoe last, plate, and banana, and \mbox{$v_{x,\max} = 0.2~\mathrm{[cm/s]}$} for the boot zipper and the shoe seam}. 

The trajectories reconstructed during contour tracking are shown in Fig.~\ref{fig:real_objects}, and the quantitative results are summarized in Table~\ref{tab:real_objects_rmse}, \sethlcolor{rev10HL}\hl{which also includes the mean and standard deviation of the control-loop execution time. These times remain close to the nominal 20~ms period for 50~Hz operation, supporting the practical real-time feasibility of the proposed implementation.}
In all cases, $r_e$ remained below $2~\mathrm{[mm]}$, the 
plate \hl{and the boot zipper exhibiting the smallest deviations, around 
$\mathrm{RMS}=0.40$ and $0.51$ $\mathrm{[mm]}$, respectively, reflecting the regularity of its geometry. 
The sensor's local deformation error $\delta_e$ is considered small, only banana showed an error in the millimeters range, the rest had errors between $12$ and $34$ times smaller.
The angular error $\alpha_e$ was highly consistent across all objects, remaining close to 
$2.7\times10^{-2}~\mathrm{[rad]}$
which indicates reliable alignment of the tactile sensor with the local contour tangent. The angular error $\beta_e$, which reflects the local curvature, showed larger variation across the evaluated contours, reaching its highest value for
the banana ($0.106~\mathrm{[rad]}$), while it is between $2$ and $4$ times smaller for the rest.}

Overall, the results reveal a clear dependence on \hl{local geometry and contact conditions of the tracked contours. Regular rigid objects, such as the plate, caused the smallest errors because their geometry simplifies feature extraction and prediction.
The zipper and seam also exhibited low errors, although these experiments were performed at a lower forward velocity due to the non-rigid nature, which favors better 
tracking conditions. 
In contrast, the banana increased the deformation and curvature related errors due to its asymmetrical and curved shape.
Consequently, the proposed VBT-MPC is capable of maintaining contact and following a variety of real-world contours, 
although the results suggest that its performance tends to deteriorate when the tactile contour is less distinguishable or when local changes in contour direction make feature extraction less reliable.}
\section{Conclusions and Future Works}
\label{sec:conclusions}
This work presented a novel Vision-Based Tactile Model Predictive Control (VBT-MPC) scheme for contour following using a markerless Vision-Based Tactile Sensor (VBTS) in an eye-in-hand configuration. Formulated as an Image-Based Tactile Servoing problem, the method used an MPC model to generate smooth end-effector motions and regulate contour-based tactile features in image space. 
Real-world experiments show that combining contour-based tactile feedback with MPC provides an effective strategy for accurate contours following with VBTS. Future work will focus on extending the method to handle moving rigid objects and incorporating explicit normal-force regulation into the control scheme, relying on forces estimated from VBTS images via learning-based models.
\bibliographystyle{IEEEtranDOI}
\bibliography{references.bib}
\end{document}